\documentclass[12pt]{article}

\usepackage{newtxtext,newtxmath}
\usepackage{comment}
\usepackage{graphicx}
\usepackage{caption}
\usepackage[letterpaper,margin=1in]{geometry}

\renewenvironment{abstract}
	{\quotation}
	{\endquotation}

\date{}

\makeatletter
\long\def\@makecaption#1#2{%
  \vskip\abovecaptionskip
  \sbox\@tempboxa{\textbf{#1} #2}%
  \ifdim \wd\@tempboxa >\hsize
    \textbf{#1} #2\par
  \else
    \global \@minipagefalse
    \hb@xt@\hsize{\hfil\box\@tempboxa\hfil}%
  \fi
  \vskip\belowcaptionskip}
\renewcommand{\fnum@figure}{Fig.~\thefigure.}
\renewcommand{\fnum@table}{Table~\thetable.}

\def\and{%                  % \begin{tabular}[t]{c}
  \end{tabular}%
  \vspace{-3pt} %\@plus.11fil%
  \begin{tabular}[t]{c}}%   % \end{tabular}

\makeatother

\usepackage{scicite}

\usepackage{url}

\def\scititle{
	Decoding Neural Population Dynamics through Robotic Analog
}
\title{\bfseries \boldmath \vspace{-10pt} \scititle}

\author{
	Wenhui~Chen$^{1\dagger}$,
	Jiyue~Tao$^{1\dagger}$,
    Yitao~Cheng$^{1}$,
    Yutong~Shi$^{1}$,
	Feitian~Zhang$^{1\ast}$,\and
    Xitong~Liang$^{2,4,5\ast}$,
    Ke~Liu$^{1,3\ast}$\and
	\small$^{1}$School of Advanced Manufacturing and Robotics, Peking University, Beijing \& 100871, China.\and  
    \small$^{2}$School of Life Sciences, Peking University, Beijing \& 100871, China.\and
    \small$^{3}$State Key Laboratory of Transvascular Implantation Devices, \and\small The Second Affiliated Hospital of Zhejiang University School of Medicine, Hangzhou \& 310009, China.\and
    \small$^{4}$IDG/McGovern Institute for Brain Research, Peking University, Beijing \& 100871, China.\and
	\small$^{5}$Peking-Tsinghua Center for Life Sciences, Peking University, Beijing \& 100871, China.\and
	\small$^\ast$Corresponding author. Email: feitian@pku.edu.cn;xitong.liang@pku.edu.cn;liuke@pku.edu.cn\and
	\small$^\dagger$These authors contributed equally to this work.
}

\begin{document} 

% Insert the title and author list
\maketitle
% Abstract, in bold
% There are strict length limits, and not all formats have abstracts.
% Consult the journal instructions to authors for details.
% Do not cite any references in the abstract.

\noindent \textbf{Teaser:} 
A robotic analog of animal neuromuscular systems reveals how neural rotational dynamics enable precise and adaptive movements.

\begin{abstract} \bfseries \boldmath 

\noindent Animal evidence shows that precise voluntary movements arise from rotational neural population dynamics in motor cortex, but their physical effects remain unknown. We developed a robotic analog of biological motor systems with artificial muscles, multimodal sensors, and a neural network controller trained via reinforcement learning. The robotic analog exhibited accurate movements, robustness to damage, and neural population dynamics akin to animals. This task-driven, embodied model illuminates the causal link between neural population dynamics and motor outcomes. We discovered that neural rotations generate oscillatory maneuvers orthogonal to the reaching direction, optimizing trajectory adjustments, which is confirmed by primate neural data. The model also revealed counterintuitive neural energy principles under sensor and motor redundancies, and striking Eureka moments during motor learning, bridging biological and artificial systems. These findings provide new perspectives on how neural dynamics contribute to accurate and flexible movement, inspiring future intelligent robots with animal-like mobility.
%\noindent Extensive evidence shows that precise voluntary movements arise from rotational dynamics of neural populations in motor cortex. However, the physical effects of such neural rotations remain unknown. Here we introduced a robotic twin as an analogous motor system, comprising artificial muscles, multi-modal sensors, and an artificial neural network, trained through reinforcement learning. The robotic twin behaved remarkably similarly to biological systems, demonstrating accurate movement, robustness against unexpected damages, and notable preparatory neural activities followed by rotational dynamics. The transparency of the robotic system enabled a lossless mapping of the causal link between neural population dynamics and motor outcomes, which is inaccessible by animal studies. We found that the rotations in the neural state space generated oscillatory maneuvers orthogonal to the reaching direction, thereby facilitating most effective trajectory adjustment. This finding was confirmed by neural data from primate motor cortex. Counterintuitively, we observed that the redundancy of sensors and motors alleviated the neural computational burden. Furthermore, the learning curve during training showed sudden emergence of rotational dynamics and accurate movements, indicating critical transitions in motor learning. These findings provide new perspectives on how neural dynamics contribute to accurate and flexible movement, inspiring future intelligent robots with animal-like mobility.
\end{abstract}
%Rotational neural dynamics in motor cortex are critical for generating precise voluntary movements, yet their mechanistic basis remain unknown. We developed a robotic twin for biological motor systems with artificial muscles, multimodal sensors, and a neural network controller trained via reinforcement learning. The robotic twin exhibited accurate movements, robustness to damage, and neural population dynamics akin to animals. This task-driven, embodied system   causally links neural population dynamics to real motor outcomes. We discovered that neural rotations generated oscillatory maneuvers orthogonal to the reaching direction, optimizing trajectory adjustments, which was confirmed by primate neural data. By bridging biological and artificial systems, this robotic twin further revealed counterintuitive neural energy principles under sensor and motor redundancies, and surprising Eureka moments during motor learning.

\subsection*{Introduction}
% The first paragraph of any Science paper does NOT have a heading
% Nor is it indented
\noindent
Humans and animals exhibit remarkable motor control, enabling both precision and flexibility in tasks ranging from agile maneuvers to delicate object manipulation \cite{todorov_direct_2000,lillicrap_preference_2013}. This capability is rooted in the motor system, which includes muscles to execute actions, sensory neurons to provide feedback, and neural systems---particularly the motor cortex---to control voluntary movement \cite{Ko2001Neuromuscular,reza_2010,Knopfel2019}. A major finding in recent decades is that the motor cortex operates through neural population dynamics. Neural population states typically evolve through an attractor-like preparatory state and transition to rotational dynamics during execution \cite{churchland2007temporal,kaufman2014cortical,churchland_neural_2012,churchland_preparatory_2024,sabatini_reach-dependent_2024,Mohammad2025}. The rotational neural dynamics are consistently observed across various tasks, including non-periodic movements. However, the exact contribution of rotation to the generation of movement remains unclear \cite{kalidindi_rotational_2021,linden_movement_2022,vyas_computation_2020}. Causally linking neural dynamics to motor output is hindered by motor system redundancy, with far more neurons than muscles in parallel pathways, and methodological limits allowing monitoring of only a small fraction of neurons \cite{barbara_2025}.

Recent advances in artificial neural network (ANN) models offer a promising bridge between neural dynamics and behavior, overcoming the experimental constraints imposed by the limited coverage of neural recordings and manipulations \cite{richards_application_2022, richards_deep_2019, bunne_how_nodate2024,Ji-An2025, yang_artificial_2020}. Convolutional neural networks (CNNs) have significantly advanced our understanding of sensory cortex, particularly the primate ventral visual stream \cite{wang_foundation_2025,bashivan_neural_2019,yamins_using_2016, cowley_mapping_2024}. Recurrent neural networks (RNNs) are employed to reflect how brains perform cognitive tasks and motor control \cite{sussillo_neural_2015,driscoll_flexible_2024}. However, most studies have focused on reconstructing simplified behavioral parameters or neuromuscular signals, often overlooking the gap between execution components and motion outcomes. Recent efforts to address this limitation have involved building biomechanical models of virtual fly or rodent \cite{aldarondo_virtual_2024,vaxenburg_whole-body_2025,chiappa2024acquiring} using physics simulators. However, the absence of real-world implementation leaves a significant gap in connecting motor output with realistic sensory information \cite{marin_vargas_task-driven_2024,vaxenburg_whole-body_2025,Wang2024Sensing,Girguis2024,Wang2024Sensing}.

To overcome the aforementioned experimental limitations and to causally link latent neural dynamics to physical movement, we propose the robotic analog, a fully analogous neuromuscular system, consisting of distributed artificial muscles and multi-modal sensors, controlled by ANNs trained via reinforcement learning (Fig.~\ref{fig:f1}A--C, fig.~S1A). The robotic analog system behaved remarkably similarly to biological motor systems, providing an embodied platform for normative, task-driven, data-free AI models to decipher movement-related neural activities. We used it to successfully decode the impact of rotational neural dynamics on movement, reveal how sensory input affects neural energy, and capture ``Eureka'' moments in motor learning.

\subsection*{Results}
% Research Articles and Reviews split the text into sections using headings
% Use a short (up 6 words) descriptive phrase, not generic 'Results' or 'Conclusions'
% Most other formats do not have headings, see the journal instructions to authors for details
\subsection*{Robotic analog of the biological motor system}
This robotic analog system is in the form of a continuum robot arm, whose actuation was achieved by the contraction of artificial muscles in response to Joule heating. The biomimetic artificial muscles were fabricated by braiding 3D printed liquid crystal elastomer (LCE) fibers with a conductive furry coating. The electric resistance of the coating varied with muscle length (Fig.~\ref{fig:f1}B), closely mimicking the proprioceptive feedback sensed by muscle spindles \cite{adem_muscle_2017,kissane2022skeletal}. Additional proprioceptive and exteroceptive sensory feedback were also incorporated into the robotic system to recapitulate the biological motor system. The proprioceptive feedback included the muscle length (``Len'') feedback from the self-sensing artificial muscles, and the joint angle (``Ang'') feedback by embedded inertial measurement units (IMUs) (fig.~S1B--H). The exteroceptive feedback was provided by the vision-based motion capture system (MoCap), measuring real-time positional information of the robot arm tip, i.e., the end effector (``Vis''). Using these sensory inputs, the ANN controller then converted the information into motor commands in the form of different stimulating currents for individual artificial muscles, through learned control policies.

In order to systematically test and optimize different variations of sensorimotor transformations, a series of ANN controllers were first trained in a simulation environment. We modeled the artificial muscles with a customized electromechanical constitutive model, integrated into a simplified skeletal model of the robot arm in the MuJoCo simulator \cite{mujoco,Li2025} (fig.~S2). The ANN controller (i.e., the actor network) has an end-to-end architecture, which receives sensory feedback and directly outputs actuator stimuli (Fig.~\ref{fig:f1}C), i.e., electric current, given the target reach point. The controller was trained using the soft actor-critic (SAC) algorithm. Both the critic and actor networks consist of one fully connected linear layer, one RNN module with 512 nodes, and another linear output layer (Fig.~\ref{fig:f1}C). RNN was adopted in both the actor and critic networks to capture temporal dependencies and receive multi-modal inputs, including vision, muscle lengths, joint angles, along with different combinations of these three inputs (fig.~S3). Considering that the motor cortex also receives input from the somatosensory cortex, where proprioceptive feedback is processed to estimate limb position \cite{marin_vargas_task-driven_2024}, we further trained a separate ANN to mimic the somatosensory cortex, predicting the spatial position of the end effector based on proprioceptive feedback (``Pred. Vis''). 

To seamlessly transfer our ANN controllers to real-world implementation, we introduced randomization during training in the simulation environment (Fig.~\ref{fig:f1}C). Surprisingly, we found that no fine-tune on the physical robot arm was required for the ANN controller. During the real-world experiment, it achieved a mean reaching error below 5~mm across all conditions (Fig.~\ref{fig:f1}D), approaching the tracking accuracy of the MoCap system (Fig.~\ref{fig:f1}E-J, fig.~S4). The trained controller generalized to arbitrary trajectories, demonstrating high-fidelity motor control with smooth movements (Fig.~\ref{fig:f1}K, Movie S1). Thus, this robotic analog system enables autonomous and adaptive movement generation, demonstrating remarkable similarity with its biological counterpart. 

% If your text is very short you might need to uncomment the following line to avoid
% layout problems with the figures and tables.
%\newpage

%%%%%%%%%%%%%%%% MAIN TEXT FIGURES %%%%%%%%%%%%%%%

\begin{center} % Do NOT use \begin{figure*}
	\includegraphics[width=0.9\textwidth]{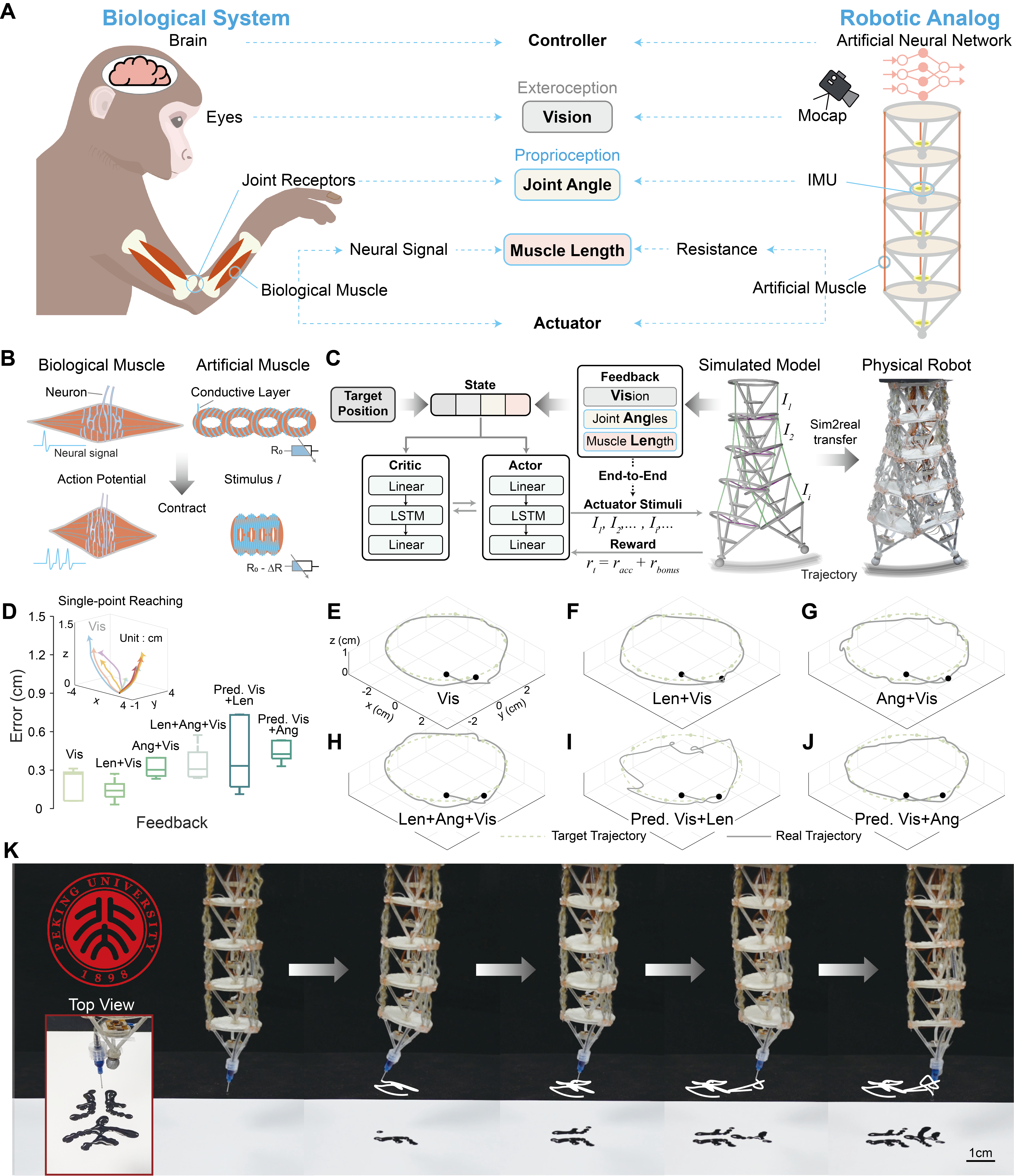} % for an image file named example_figure.*
	% Pick an appropriate width - in print, figures are usually one or two columns wide, which can
	% be approximated by 0.3\textwidth or 0.6\textwidth respectively. Use appropriate label sizes.

	% Captions go below figures
    
    % \renewcommand{\baselinestretch}{0.8}\selectfont
    \captionof{figure}{\textbf{Design, sensing and control of the robotic analog.}
    \small (\textbf{A}) Conceptual schematic comparing the biological motor system and the robotic analog. The biological system integrates multiple feedback signals (e.g., vision, joint angle, muscle length) to control muscle actuation via the brain; the robotic analog replicates this structure through artificial muscles, proprioceptive sensing, and visual feedback, enabling closed-loop control via an ANN controller. (\textbf{B}) Self-sensing mechanism of biological muscle and artificial muscle. (\textbf{C}) Overview of the end-to-end reinforcement learning framework. The actor network receives multi-modal input and directly outputs stimulating currents to the artificial muscles. (\textbf{D}) Reaching error of ANN controllers receiving different feedback combinations. The inset shows the single-point reaching trajectories of the robot arm under vision feedback. (\textbf{E--J}) Circular trajectory tracing of models with different feedback combinations: (\textbf{E}) vision only (``Vis''), (\textbf{F}) muscle length + vision (``Len+Vis''), (\textbf{G}) joint angle + vision (``Ang+Vis''), and (\textbf{H}) muscle length + joint angle + vision (``Len+Ang+Vis''); (\textbf{I--J})  proprioception-only models: (\textbf{I}) predicted vision + muscle length (``Pred. Vis+Len'') and (\textbf{J}) predicted vision + joint angle (``Pred. Vis+Ang''). (\textbf{K}) Demonstration of the robot arm tracing complex trajectories by writing Peking University's emblem within 2 minutes.}
\label{fig:f1} % give each figure a logical label name
\end{center}

\subsection*{Similarities between robotic and biological motor control}
The biomimetic structure of the robotic analog system yielded highly similar behavior to the biological motor system. Firstly, we found that the robotic analog system exhibited emergent robustness to unexpected damages, maintaining accurate movements without requiring retraining. Three representative failure modes were tested: failure of a single muscle, failure of two muscles in different layers, and failure of three muscles within the same layer (Fig.~\ref{fig:f2}A). Without any pretraining or adjustment, the robot arm autonomously adapted to such unseen scenarios by leveraging other functional muscles to compensate for reaching accuracy, kept consistently below 5~mm in real-world experiments (Fig.~\ref{fig:f2}B, fig.~S5). 

 A striking emergent feature of the robotic analog was observed in the neural dynamics. Without any pre-determined task structure, while the kinematic trajectories (Fig.~\ref{fig:f2}C) showed the physical task execution, the population activity of 512 RNN neurons (Fig.~\ref{fig:f2}D) exhibited two distinct phases during various tasks recapitulating the separation of preparatory state and execution state observed in motor cortex \cite{kaufman2014cortical,churchland_preparatory_2024}. In motor cortex, preparatory activity is largely confined to an output-null subspace, functionally decoupled from execution-related activity. Remarkably, a similar structural decoupling emerged in the robotic analog system at the subspace level. To characterize the relationship between preparatory and execution activity, we first examined the leading principal components, which capture the dominant variance direction and are closely related to movement output. The angle between the first PCs (explaining 20.39+3.10\% of the variance) is close to $90^\circ$(Fig.~\ref{fig:f2}E), indicating that their dominant directions are nearly orthogonal. We further quantified the geometric relationship between the two manifolds by computing the full set of principal angles between their subspaces (fig. S6A). The angles are broadly distributed, with most values between $30^\circ$ and $90^\circ$, suggesting that the separation between preparatory and execution activity extends across multiple dimensions rather than being confined to a single axis. 

While this geometric analysis suggests a separation between the two subspaces, it does not quantify how neural activity is distributed across them. To address this, we performed a variance-based analysis and found that a substantial fraction of execution variance lies in the null space of the preparatory subspace, and conversely, a large portion of preparatory variance lies in the null space of the execution subspace (both $>$0.5 across different feedback combinations; fig.~S6B), while the subspace overlap remains markedly lower than 0.5. These results indicate that preparatory and execution activity are largely segregated, yet not strictly orthogonal, forming distinct but partially overlapping neural manifolds.

Consistent with this subspace organization, the variance of neural population activity showed a sharp reduction at the transition between the two states (Fig.~\ref{fig:f2}F), mirroring the attractor-like neural dynamics in the preparatory subspace of motor cortical neurons, of which activity converges to set initial states for the subsequent execution states \cite{Churchland2010}. Similar to the motor cortex, the subspace of the robotic analog's preparatory state was shown to be generally output-null, ensuring that ongoing muscle activity remains unaffected while still specifying the upcoming movement (Movie S2).

To further understand the dynamical mechanism underlying this transition, we performed a local linearization analysis of the network dynamics \cite{michaels2020goal,Michaels2016}. Despite the nonlinear mapping between neural activity and behavior, a linear dynamical approximation accurately captured the evolution of neural states (R$^2$ : 0.9–1.0; fig.~S6C), indicating that the system operates in a locally linear regime. Within this framework, neural dynamics during the preparatory phase are dominated by activity along output-null directions, whereas during execution, the dynamical flow is progressively reoriented toward output-potent directions (fig.~S6D-E), reflecting a redistribution of neural dynamics rather than a global increase in gain.

We also performed a mode-based analysis of the linearized dynamics, which showed that behavioral performance is dominated by a small number of dynamical modes, as removing the top-ranked modes leads to substantial performance degradation (fig. S6F). These dominant modes are not strongly aligned with output-potent directions, suggesting that behaviorally relevant computations are primarily carried out within internal dynamical structure before being transformed into output-relevant signals.

Next, we examined how these neural subspaces relate to behavior. Using canonical correlation analysis (CCA), we quantified the alignment between neural population activity and kinematics\cite{Gallego2018,Dekleva_2025}. Both preparatory and execution activity exhibit low-dimensional neural–kinematic alignment, with correlations concentrated in a small number of leading modes (fig. S7). This alignment depends on behavioral epoch, as mappings derived from one epoch do not generalize to the other (fig. S8). Comparisons with tensor maximum entropy (TME) surrogate data show that these structures are not reproduced by surrogate datasets and reflect population-level organization.

In addition to subspace reorganization, we observed a significant displacement of neural population activity between preparatory and execution states in the shared state space (fig.~S9), indicating a global shift between the two manifolds.

More importantly, during the execution state, we found that the neural population activity of ANN controllers exhibited rotational dynamics. In primate motor cortex, such rotational dynamics is ubiquitously observed across various tasks, including single-point reaching and cycling \cite{churchland_neural_2012,Russo2020Neural}. The rotational dynamics reveal how the motor cortex generates patterns of muscle activity \cite{Vyas2020Computation}. In this deterministic dynamical system, where future activity depends on past population activity, the phase and amplitude of neural rotations are shaped by the preparatory activity that sets the initial conditions for the subsequent execution dynamics (Fig.~\ref{fig:f2}G).
In the robotic analog system, we observed characteristic spiral-like trajectories in the population activity of the 512 RNN neurons (Fig.~\ref{fig:f2}H). To capture the rotational component, we applied the jPCA dimensionality reduction method \cite{churchland_neural_2012}, which reveals latent rotational structures by fitting a skew-symmetric matrix to the temporal derivatives of neural population activity. The resulting jPCs form an orthonormal basis spanning the same subspace, with the first jPC plane capturing the strongest rotational tendency. We observed consistent rotational dynamics in both single-point reaching and cycling tasks (Fig.~\ref{fig:f2}I--J) and across all variations of ANN controllers with different feedback combinations (Fig.~\ref{fig:f2}K--L, fig.~S10A).

\begin{center} % Do NOT use \begin{figure*}
	\includegraphics[width=1\textwidth]{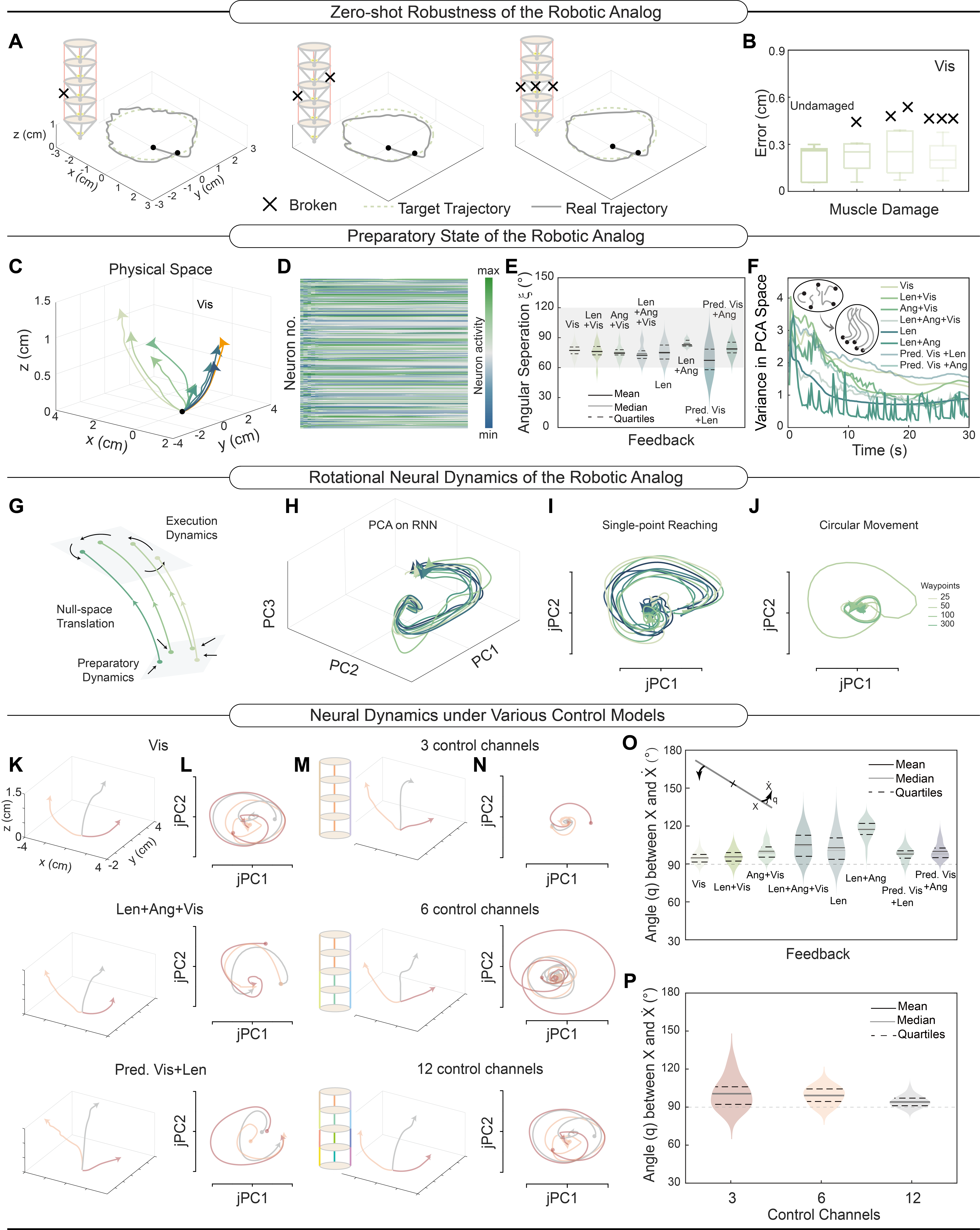} % for an image file named example_figure.*
	% Pick an appropriate width - in print, figures are usually one or two columns wide, which can
	% be approximated by 0.3\textwidth or 0.6\textwidth respectively. Use appropriate label sizes.

	% Captions go below figures
    %\renewcommand{\baselinestretch}{0.8}\selectfont
	\captionof{figure}{\textbf{Functional similarity between the robotic analog and the biological motor system.}
		\small (\textbf{A}) The robot arm maintained accurate trajectories despite various unexpected artificial muscle damages. (\textbf{B}) Reaching errors in different damaged scenarios with visual feedback (``Vis''). (\textbf{C}) Representative single-point reaching trajectories in 3D movement space with visual feedback. (\textbf{D}) Neural activities in  the actor network (RNN layer) during reaching (orange trajectory in \textbf{C}). (\textbf{E})  Distribution of the angular separation ($\xi$) between the PC1 in the first 1.2s and the subsequent time period, under different feedback combinations. (\textbf{F}) Temporal evolution of variance in the PCA space, by projecting the original 512 neural dimensions onto the first 30 PCs. The inset illustrates initially dispersed neural activity becoming concentrated over time. (\textbf{G}) Schematic illustration of the dynamics in the neural state space at different stages of movement. (\textbf{H}) Neural activities projected onto the first three PCs of the actor network, visualized to reveal the underlying rotational structure. (\textbf{I}) Projection of the neural population response of the actor network (512 neurons, 10 single-point reaches, ``Vis'') into the two dimensional jPCA space. Each trace shows the first 10s of movement-related activity departing from the preparatory state. The units are arbitrary. (\textbf{J}) Projection of the actor network activity (512 neurons, 4 circular paths, ``Vis'') into the two dimensional jPCA space during circular movements. Each trace plots the first 40s of movement-related activity away from the preparatory state. (\textbf{K}) Reaching trajectories of the robot arm under different feedback combinations. (\textbf{L}) The jPCA projections of the actor network activity under different feedback combinations during single-point reaching (as in \textbf{I}). (\textbf{M}) Schematics and trajectories of robot arms using 3, 6, and 12 control channels. (\textbf{N}) The jPCA projection of the actor network activity (512 neurons, ``Vis'') under different control channels during single-point reaching (as in \textbf{I}). (\textbf{O}) Distribution of the angle $q$ between the neural state $x$ and its derivative $\dot{x}$. The angle was measured as illustrated schematically (inset) after projecting the data onto the first jPCA plane. (\textbf{P}) Distribution of $q$ across different numbers of control channels.}
	\label{fig:f2} % give each figure a logical label name
\end{center}

 To quantitatively assess rotational dynamics in our robotic analog system, we measured the alignment ($q$) between neural activity and its temporal evolution, together with the goodness of fit ($R^2$) of a low-dimensional rotational model. In both vision-driven and proprioception-driven models, we observed that $q$ consistently approached $90^\circ$ (Fig.~\ref{fig:f2}O), with $R^2$ values significantly exceeding those derived from TME-based surrogate datasets that preserved the covariance structure of the original data (fig.~S10B). Such rotational dynamics were observed exclusively in the RNN layers and not in the input linear layer and the muscle output signals (fig.~S10A--D)

Our models incorporating RNN layers, trained under identical task conditions and across multiple random seeds, were compared with a feedforward multilayer perceptron (MLP) model. While our model consistently exhibited strong rotational structure in the RNN layers, MLP models showed substantially weaker and less coherent rotations, together with degraded behavioral performance (fig. S11A–B). These differences were robust across random initializations, indicating that the observed rotational dynamics are not a consequence of a particular seed or parameter configuration. 

We further investigated how rotational dynamics is affected by motor redundancy. The system exhibits motor redundancy when the number of control channels exceeds three, as the end effector has 3 degrees of freedom (Movie S3). When comparing systems with 3, 6, and 12 independent control channels, we found that increasing the number of control channels led to progressive improvements in both the consistency of rotational neural dynamics and movement accuracy (Fig.~\ref{fig:f2}M--N). The 12-channel configuration exhibited the most concentrated angular distributions around $90^\circ$, indicating that neural population activity strictly followed rotational dynamics, thereby enabling the most effective movement control (Fig.~\ref{fig:f2}P, fig.~S12). We noted that rotational dynamics were consistent in both simulations and real-world experiments (fig.~S10, S12), indicating sensory feedback delays and noise do not influence neural rotations. Collectively, these findings further underscore the critical role of rotational neural dynamics in motor control.

\subsection*{Decoding rotational neural dynamics}
Rotational dynamics are widely recognized as a characteristic feature of motor cortex population activity, while their precise functional role, particularly their relationship to movements, remains an active area of research\cite{kalidindi_rotational_2021}. To decode the role of rotational dynamics in motor control, we leveraged the full transparency and accessibility of our robotic system. In this system, we can construct a lossless mapping from the neural state space to the muscle action space, and then to the physical motion space (Fig.~\ref{fig:f3}A). From the orthonormal basis vectors of the first jPC plane, we selected the first jPC axis representing the strongest neural oscillation and reconstructed the RNN neural signal through PCA reconstruction. This transient RNN signal was used repeatedly to generate a 12-dimensional action signal for the artificial muscles, driving the robot arm to produce the resultant movement of the end effector in the physical movement space, termed ``jPC1-reconstructed'' (Fig.~\ref{fig:f3}A). We found that the ``jPC1-reconstructed'' movement was consistently orthogonal to the reaching direction, with the angle $\theta$ between the ``jPC1-reconstructed'' and the reaching direction staying near $90^\circ$ (Fig.~\ref{fig:f3}B--C). This result suggests that rotational neural dynamics directly drive movements orthogonal to the target direction. We hypothesized that such dynamics primarily facilitate trajectory adjustments as orthogonal maneuvers are the most effective in correcting the reaching direction.

\begin{center} % Do NOT use \begin{figure*}
	\includegraphics[width=1\textwidth]{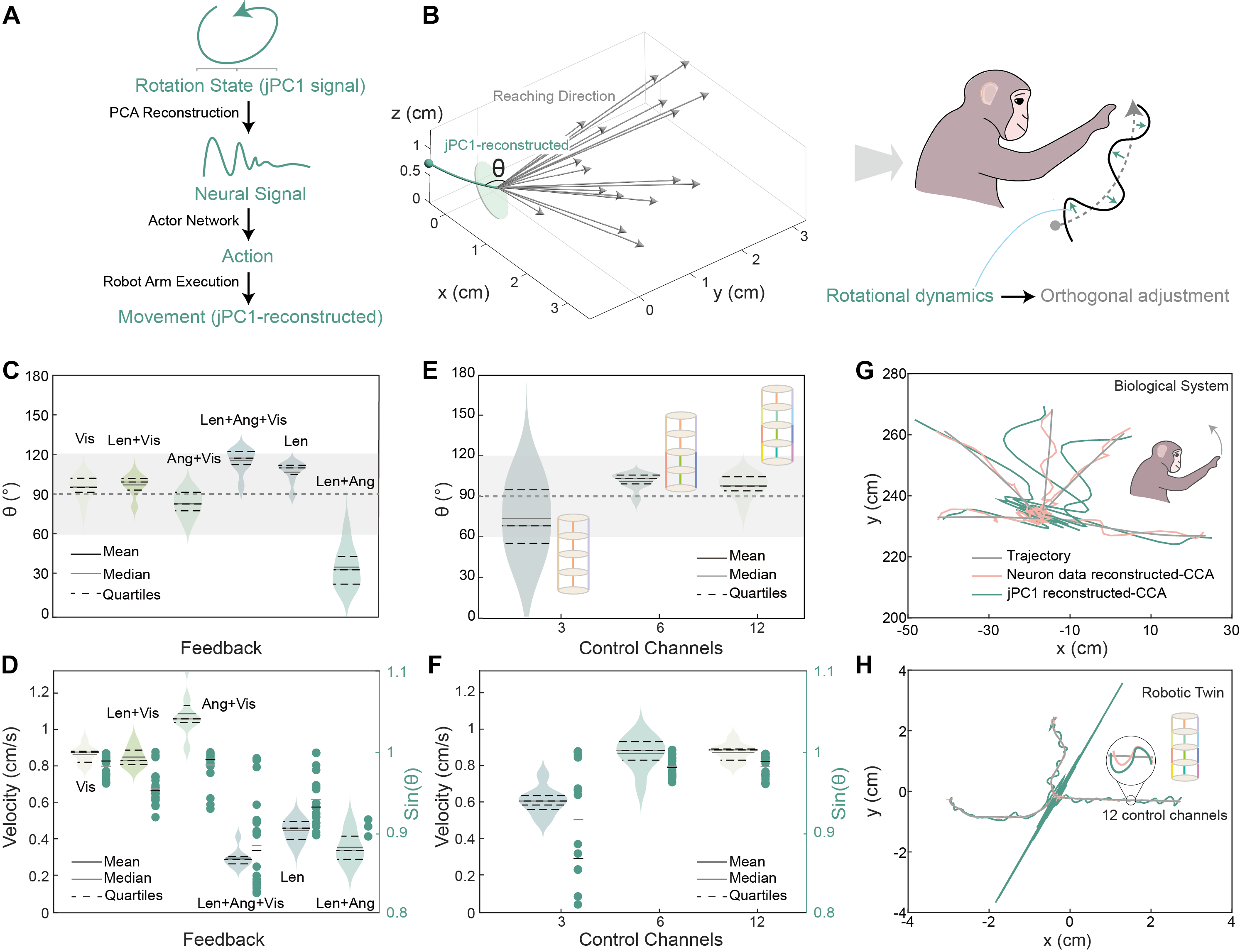} % for an image file named example_figure.*
	% Pick an appropriate width - in print, figures are usually one or two columns wide, which can
	% be approximated by 0.3\textwidth or 0.6\textwidth respectively. Use appropriate label sizes.

	% Captions go below figures
    %\renewcommand{\baselinestretch}{0.8}\selectfont
	\captionof{figure}{\textbf{Physical effect of rotational neural dynamics on movement.}
		\small (\textbf{A}) Schematic illustration of the extraction process of the ``jPC1-reconstructed'' movement, representing movement dominated by rotational dynamics. (\textbf{B}) The distribution of the ``jPC1-reconstructed'' movement relative to the direction of the single-point reaching task (reaching direction), by the ``Vis'' model. The right panel shows the physical implication of the rotational dynamics. (\textbf{C}) The distribution of the angle ($\theta$) between ``jPC1-reconstructed'' movement and the reaching direction under different feedback combinations. (\textbf{D}) The distribution of $sin(\theta)$ and velocity during single point reaching tasks under different feedback combinations. (\textbf{E}) The distribution of the angle $\theta$ between ``jPC1-reconstructed'' movement and the reaching direction with different control channels. (\textbf{F}) The distribution of $sin(\theta)$ and velocity during single point reaching task with different control channels. (\textbf{G}) The actual reaching trajectories of the monkey, compared with the CCA projections of the neural population response and the reconstructed jPC1 signal. (\textbf{H}) The actual reaching trajectories of the robot arm, compared with the CCA projections of the neural population response and the reconstructed jPC1 signal. The ``Vis'' model is used with 12 control channels.}
	\label{fig:f3} % give each figure a logical label name
\end{center}

The angle $\theta$ exhibited a consistent tendency towards $90^\circ$ across all variations of ANN controllers with different feedback combinations (Fig.~\ref{fig:f3}C). Among these variations, we observed that as the ``jPC1-reconstructed'' movement approached perfect orthogonality to the reaching direction ($\sin(\theta)$ nearing 1), the controllers achieved higher corresponding movement velocity (Fig.~\ref{fig:f3}D). We further analyzed variations of ANN controllers with different levels of control redundancy. In the robotic system with three independent control channels, and thus without motor redundancy, the angle between the ``jPC1-reconstructed'' movement and the reaching direction exhibited a substantial deviation from $90^\circ$ (Fig.~\ref{fig:f3}E). This suggested that the lack of control redundancy hinders the decoupling of muscle forces into orthogonal directions. In contrast, increasing the number of independent actuators (6, 12 control channels), thereby increasing control redundancy, led the $\theta$ angle to converge toward $90^\circ$, while simultaneously improving movement velocity and accuracy (Fig.~\ref{fig:f3}F). Collectively, our systematic comparison of various input-output configurations supports the idea that rotational neural dynamics drive orthogonal maneuvers and facilitate efficient, accurate movement execution.

To verify whether the mechanistic role of rotational neural dynamics in leading orthogonal adjustment of trajectory demonstrated in the robotic system also applies to biological systems, we tested this hypothesis using neural recording data from the primate primary motor cortex (M1, data from \cite{kalidindi_rotational_2021}). To directly compare neural activity with movement trajectory, we used canonical correlation analysis (CCA) \cite{thompson2000cca}, which finds a linear mapping that maximizes correlation between neural space and movement space. The CCA mapping of population activity from 141 M1 neurons yielded reconstructed movements (Fig.~\ref{fig:f3}G, ``Neuron data reconstructed-CCA'', pink lines) that closely aligned with the monkey’s actual reaching trajectories (gray lines). Using the same linear mapping, we reconstructed movements driven by the jPC1 signal (``jPC1 reconstructed-CCA'', green lines), which showed pronounced orthogonal oscillations along the actual trajectories (Fig.~\ref{fig:f3}G). Here, jPCA is used to extract rotational components of neural activity rather than to perform decoding, enabling us to isolate their contribution to movement structure under the same neural–kinematic mapping.

This observation strongly suggested that in motor cortex, the rotational component of neural signals is also associated with orthogonal adjustment of reaching direction. We further applied the same CCA method to the robotic analog systems and observed the same phenomenon across all models with various sensory feedback combinations and control strategies (Fig.~\ref{fig:f3}H). In addition, the amplitude of this jPC1-driven oscillation was directly linked to motor performance (Fig.\ref{fig:f2}K, fig.~S12A): the functional ``Pred. Vis+Len'' model displays significantly greater oscillation amplitudes than the non-functional ``Len'' model, which failed to reach the target. This cross-domain validation strongly supported the functional role of rotational neural dynamics in driving the orthogonal maneuvers and trajectory adjustment. These observations naturally raise the question of whether the rotational component directly contributes to target-directed movement, or instead reflects a distinct dynamical process that is functionally separated from the target-driving signals.

To investigate this possible separation, we first decomposed the population activity into an output-potent (target-driving) subspace and a rotational (jPC) subspace. The output-potent subspace captures most of the target-related signal (R$^2$: 0.8–0.9), whereas the jPC subspace explains only a small fraction of the variance (R$^2$ $<$ 0.15). The two subspaces are geometrically distinct, with principal angles between 60\textdegree and 80\textdegree; fig.~S13A–B). These results suggest that target-directed control is primarily mediated by the output-potent subspace, whereas rotational dynamics occupy a distinct component of neural activity that support orthogonal trajectory adjustment and movement stabilization.

Despite this geometric separation, targeted neural ablation revealed that the rotational component remained functionally important for accurate motor performance. Selectively pruning the output connections of units that drive the rotational subspace increased reaching error (fig. S13C-E). This degradation was greater than in either intact networks or randomly pruned networks, suggesting that rotation-dominant units contribute to accurate movement execution.

Together, our findings elucidate the mechanistic role of rotational neural dynamics in motor control and revealed a universal control principle shared by both biological and artificial systems \cite{Mohammad2025}.

\begin{center} % Do NOT use \begin{figure*}
	\includegraphics[width=1\textwidth]{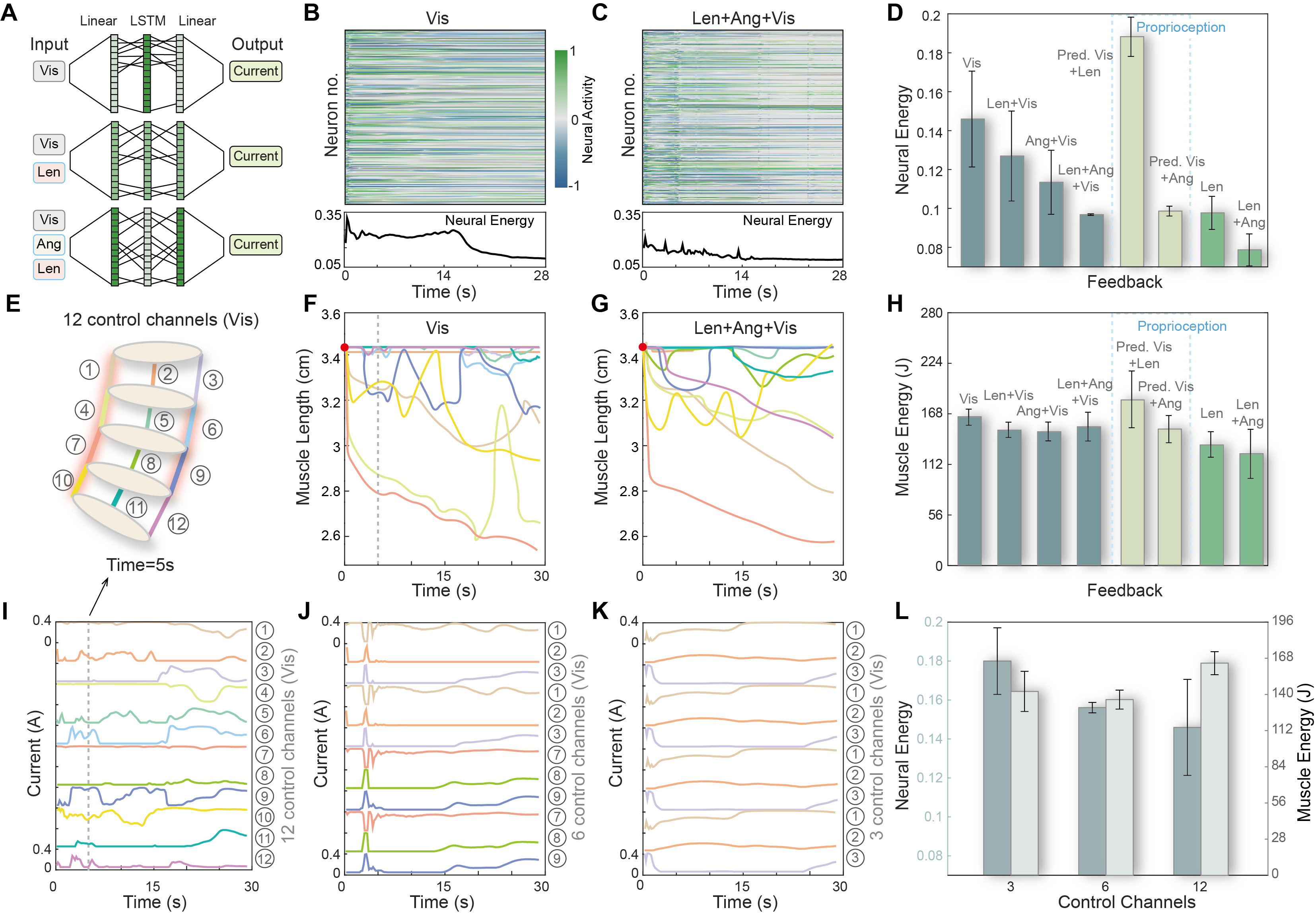} % for an image file named example_figure.*
	% Pick an appropriate width - in print, figures are usually one or two columns wide, which can
	% be approximated by 0.3\textwidth or 0.6\textwidth respectively. Use appropriate label sizes.

	% Captions go below figures
    %\renewcommand{\baselinestretch}{0.8}\selectfont
	\captionof{figure}{\textbf{Effect of sensory feedback and motor redundancy on neural and muscle energy.}
		\small (\textbf{A}) Schematic of the actor networks with different sensory inputs. Each network consists of an input linear layer, an RNN layer (LSTM), and an output linear layer to control motor current. (\textbf{B} and \textbf{C}) Neural activities of the ``Vis'' model and the ``Len+Ang+Vis'' model driving the robot arm to reach the same point ($x$,$y$,$z$)=(3, 0, 1). The subplots below \textbf{B} and \textbf{C} show the corresponding total neural energy, calculated as the squared sum of all neural activities. (\textbf{D}) Average neural energy over time of the RNN layer in actor network under different feedback combinations. (\textbf{E}) Muscle activation pattern of the robot arm at 5 second during reaching with 12 control channels by the ``Vis'' model. (\textbf{F} and \textbf{G}) Muscle lengths over time of the robot arm by  the ``Vis'' model and the ``Len+Ang+Vis'' model, respectively. (\textbf{H}) Total muscle energy, calculated as the work done by the electric currents, of the robot arm under different feedback combinations. (\textbf{I--K}) Corresponding stimulating currents for the artificial muscles during reaching with \textbf{I}-12, \textbf{J}-6,and \textbf{K}-3 control channels. (\textbf{L}) Neural and muscle energy of the robot arm with different control channels.}
	\label{fig:f4} % give each figure a logical label name
\end{center}

\subsection*{Neural energy under different feedback and actuation combinations}
We then characterized the neural population dynamics from the perspective of neuron energy. Here, energy is interpreted as a measure of control and computational efficiency, linking neural population dynamics to behavioral output. Its variation tracks changes in both neural activity structure and motor performance, providing a unified perspective on how neural dynamics support efficient control. By comparing ANN controllers with different sensory feedback combinations, we observed a counterintuitive pattern in neural energy, defined as the time-averaged squared activity of all neurons during each single-point reaching task. While it is intuitive to expect that processing more sensory inputs would increase overall neural energy, this was only observed in the input linear layers. In contrast, the RNN layers showed reduced energy with more diverse sensory feedback (Fig.~\ref{fig:f4}A). As sensory channels increased from ``Vis'' only to ``Len+Ang+Vis'', RNN activity converged more rapidly to a stable low-energy state, resulting in reduced overall neural energy per movement (Fig.~\ref{fig:f4}B--C, fig.~S14. This pattern was consistently observed in both simulated and physical robot arms, with RNN energy decreasing and linear layer energy increasing as sensory feedback channels increased (Fig.~\ref{fig:f4}D, fig.~S14I,M). 

While increasing sensory feedback channels reduced overall RNN neural energy (Fig.~\ref{fig:f4}D), the total muscle energy quantified by summed electrical works applied to the artificial muscles per movement, remained approximately constant (Fig.~\ref{fig:f4}E--H, fig.~S14J--K).Nonetheless, distinct muscle activation patterns emerged across feedback combinations (Fig.~\ref{fig:f4}E--G). 
To assess the robustness of these coordination patterns, we trained the SAC controller across multiple random seeds (n = 15) for each feedback configuration.While individual muscle activation profiles varied in detail, particularly in terms of which units underwent the greatest deformation, the structural trends across feedback conditions remained consistent (fig. S15). Across all seeds, the vision-only (‘Vis’) model produced sparse control strategies dominated by a limited subset of muscles. In contrast, integrating proprioceptive feedback (‘Len+Ang+Vis’) shifted the controller toward more distributed and smoother activation profiles. Across seeds and feedback conditions, distal muscles tended to exhibit earlier and stronger activation than proximal ones. This pattern was preserved despite variability in individual muscle contributions. These observations indicate that increased sensory feedback does not impose a fixed actuation pattern, but is associated with a more distributed use of more involved actuation pattern.

We further investigated the effect of control redundancy on the neural energy and muscle energy (Fig.~\ref{fig:f4}I--K). Similar to adding sensory channels, increasing control channels reduced neural energy, while maintaining comparable muscle energy (Fig.~\ref{fig:f4}L). It also led to more frequent and variable actuation adjustments, indicating greater control flexibility. These patterns were consistently observed in both simulation and physical experiments (fig.~148I--K, P--Q). Together, these findings demonstrate that both sensory feedback and motor redundancy improve neural computation efficiency without increasing actuation energy demand.

\subsection*{Eureka moments during the learning process}
A fundamental limitation in biological motor system studies is the difficulty of tracking the full course of neural development, especially in primates. Our robotic analog system overcomes this limitation by unrolling the training process of the ANN controllers, uncovering how neural networks develop motor control capabilities. The ANN controllers were trained by the Soft Actor-Critic (SAC) algorithm, an off-policy RL method that iteratively optimizes action policies over repeated episodes (Fig.~\ref{fig:f5}A). In each episode, once the end effector successfully reached the current target within a tolerance distance (``hit''), a new target position was randomly assigned. The robot arm was rewarded for reaching as many targets as possible within a fixed time window. If it failed to reach the target (``miss''), it continued attempting until time expired, then reset for the next episode. After training in the simulation environment, ANN controllers were deployed to control the physical robot arm in real-world experiments (Fig.~\ref{fig:f5}B). Models that incorporated exteroceptive feedback transferred directly without any further training. For those models relying solely on proprioception, the controller network required no retraining, while the network predicting the spatial position of the end effector (``Pred. Vis'') based on proprioceptive feedback required minimal real-world fine-tuning.

\begin{center} % Do NOT use \begin{figure*}
	\includegraphics[width=1\textwidth]{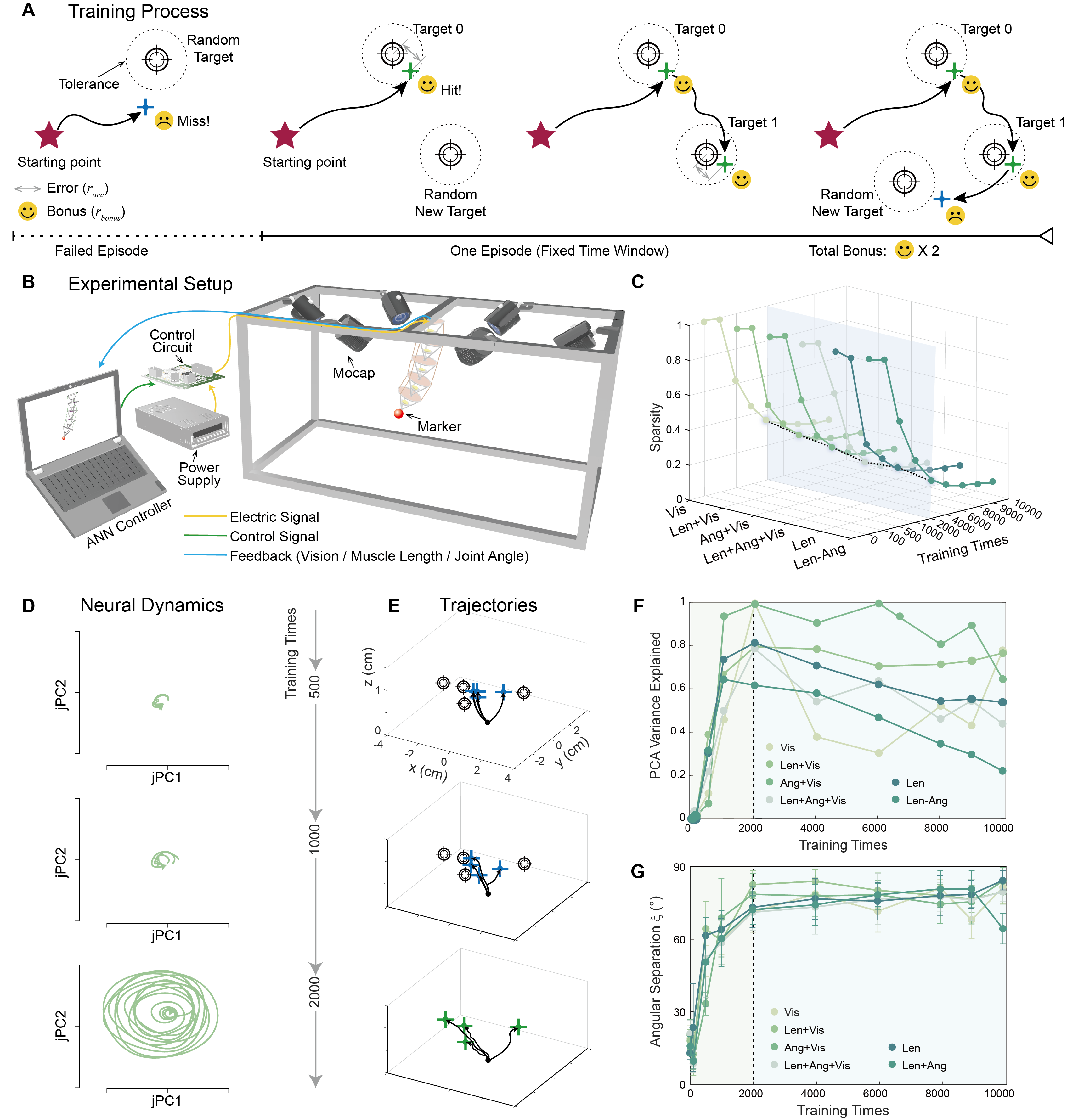} % for an image file named example_figure.*
	% Pick an appropriate width - in print, figures are usually one or two columns wide, which can
	% be approximated by 0.3\textwidth or 0.6\textwidth respectively. Use appropriate label sizes.

	% Captions go below figures
    %\renewcommand{\baselinestretch}{0.8}\selectfont
	\captionof{figure}{\textbf{Changes of neural population dynamics and motor performance during the learning process.}
		\small (\textbf{A}) Training protocol of the robot arm, which is driven to achieve optimal accuracy, as well as collecting most bonuses in a given time window. (\textbf{B}) The experimental setup of the robot arm. Six motion capture cameras are mounted on a custom metal frame to track the spatial coordinates of the marker on the end effector. Muscle length data is collected by the control circuit. Joint angle data is directly received by the computer via the serial port. The actor network generates control signals that are converted to 12 electric current outputs for the artificial muscles. (\textbf{C}) Network sparsity of actor network with different feedback combinations during the learning process. The training time refer to the count of training episodes. (\textbf{D}) the jPCA projection of the actor network activity (``Vis'' model) at different training episodes during single-point reaching. Each trace plots the first 10s of movement-related activity away from the preparatory state. (\textbf{E}) Movement trajectories of the robot arm at different training episodes, showing a sudden shift from inaccurate reaching in the early stage to precise reaching near episode 2000. (\textbf{F}) Variance explained by the first two PCs (out of 30) of the actor network activity under different feedback combinations during learning. (\textbf{G}) The angular separation ($\xi$) between the PCA subspace in the first 1.2s and the subspace in the subsequent time period, computed as the angle between their corresponding first PC, under different feedback combinations.}
	\label{fig:f5} % give each figure a logical label name
\end{center}

By comparing ANN controller models at different training stages, we observed sudden transitions among multiple aspects of network features. First, the sparsity of neural representations showed a sharp decline and stabilized near the 2000th training episode across all models with different feedback combinations (Fig.~\ref{fig:f5}C, Movie S4). Second, stable rotational neural dynamics only emerged after 2000 episodes of training (Fig.~\ref{fig:f5}D, fig.~S16), accompanied by a dramatic jump and subsequent plateaus in PCA-explained variance and rotation radius (Fig.~\ref{fig:f5}F, fig.~S17A). Third, at the same point, after 2000 training episodes, the robot arm suddenly began achieving accurate reaching of target positions, followed by slower behavioral performance improvement afterwards (Fig.~\ref{fig:f5}E, fig.~S17B). Fourth, the separation between preparatory and execution states emerged at the same time stamp, showcased by a sudden shift in the angular separation $\xi$ of two subspaces towards $90^\circ$, with the preparatory state becoming orthogonal to the execution state (Fig.~\ref{fig:f5}G). Lastly, sudden transitions in neural energy and muscle actuation energy were also observed at 2000th training episode (fig.~S17C-D).

To assess the robustness of this transition, we evaluated multiple architectures, sensory feedback combinations, and random initializations (fig. S18–S19). Across all tested settings, a similar transition was observed, accompanied by concurrent changes in neural sparsity, population dynamics, and behavioral performance. While the exact transition point varied slightly across random seeds and configurations, the overall pattern of coordinated change was preserved. Both RNN and MLP architectures exhibited such transitions, although the RNN layers developed stronger rotational dynamics and achieved higher performance.

This transition reflects the emergence of a stable control regime in a closed-loop system and suggests a possible mechanism for stage-like changes in learning, arising from the interaction between neural dynamics and body mechanics \cite{w._2013,laura_2022}. However, the detailed neural mechanisms underlying this reorganization remain to be investigated.

\subsection*{Discussion}
Neural population dynamics provide a mechanistic framework for understanding how the brain performs motor control, representing one of the most influential conceptual shifts in neuroscience. However, this framework lacks definitive approaches to determine causal relationships between population-level properties and behavioral outcomes \cite{churchland_preparatory_2024,zhong_unsupervised_2025,aldarondo_virtual_2024,Mathis2024Decoding,Xiangxiao2025}. To bridge this gap, we introduced a robotic analog, an artificial neuromuscular system comprising artificial muscles, multi-modal sensors, and an ANN controller trained via reinforcement learning. While achieving motor performance comparable to biological systems, with accurate movement and robustness, the robotic analog system most notably exhibits the rotational neural dynamics characteristic of the biological motor cortex. Serving as a new paradigm that shifts beyond \textit{in vivo} and \textit{in silico} methods, the robotic analog approach provides full transparency and manipulability to causally link latent neural dynamics to behavioral relevance. Using this approach, we decoded how rotational neural dynamics shape trajectories of reach , revealed neural energy principles under various sensory inputs, and captured Eureka moments during the learning process. These observations reveal fundamental computational principles underlying robust motor control shared by biological and artificial systems.

In the motor cortex, the rise and function of rotational neural dynamics have been debated, with explanations ranging from the recurrent cortical connections to sensory feedback delays \cite{sabatini_reach-dependent_2024, kalidindi_rotational_2021,linden_movement_2022}. Using the robotic analog, we demonstrated that the dominant component of rotational neural dynamics drives orthogonal maneuvers for online correction of moving direction, enabling the most effective trajectory adjustments that compensate for errors while maintaining stability. The separation of preparatory and execution subspaces, emerging without any pre-designed structure (Fig.~\ref{fig:f2}C--H) and persisting across real and simulation settings regardless of feedback delays (Fig.~\ref{fig:f2} and fig.~S6), underscores that these dynamics arise to serve specific functional purposes rather than being driven by sensory delays. The strong coupling between behavioral performance and neural rotations across various input-output configurations (Fig.~\ref{fig:f2}K--P, Fig.~\ref{fig:f3}C--F) and during the learning process (Fig.~\ref{fig:f5}D--G)  further supports their functional relevance, which is additionally corroborated by causal perturbation analyses. This discovery also aligns with findings in the cerebellum, where orthogonal Purkinje cell vectors suppress unwanted motor output for precise behavior \cite{Mohammad2025}, suggesting a general principle of motor control.

The robotic analog offers a unique platform to study how motor systems exploit sensory and motor redundancy by varying input and output channels---an approach not feasible in biological systems. Through comparisons of neural and muscle energy across various control strategies and feedback combinations, we found that enriching feedback and increasing control channels can reduce the computational burden on the neural controller, while maintaining motor performance (Fig.~\ref{fig:f4}). Rich inputs and outputs enhance the disambiguation of sensorimotor mappings, enabling the neural controller to operate with greater efficiency. From an optimization perspective, augmenting neural state space and action space representations provides more redundant choices to the neural controller, thereby increasing the chance of identifying optimal solutions (not necessarily unique) for state estimation and motor planning. These findings highlight the value of multi-sensory integration and control redundancy for enhancing both robustness and energy efficiency \cite{changyu_2025,kenway_2012,Knopfel2019}.

While Eureka moments are typically studied in cognitive tasks, our embodied system revealed analogous insight-like transitions during motor learning. These transitions allow us to further decode their functional role in reorganizing neural control policies. Specifically, at the 2000th training episodes, we observed a sudden simultaneous emergence of rotational neural dynamics, preparatory states, and accurate movements (Fig.~\ref{fig:f5}, fig.~S10). This transition was accompanied by an abrupt drop in the sparsity of network activations, indicating a critical transition of the internal representations of the neural network. Importantly, this transition is consistently observed across different random seeds, architectures, and sensory feedback combinations, suggesting that it reflects a robust feature of the learning dynamics rather than a configuration-specific effect. This insight-like reorganization echoes prior findings in biological motor learning, where neural population activity becomes increasingly correlated and stable as skill develops, reflecting a shift from exploratory, high-variance patterns to coordinated, expert-level dynamics \cite{Peters2014Emergence}.  

As neuroscience progresses toward mechanistic and computationally grounded explanations of neural population dynamics, our work directly addresses the growing need for platforms that causally bridge the neural state space and behavioral space. By integrating physical embodiment with computational transparency, the robotic analog enables hypothesis-driven experiments and fine-grained analysis that would be challenging or infeasible for \textit{in vivo} experiments. Therefore, the robotic analog provides a generalizable paradigm to investigate the principles shaping motor control and behavior.

Beyond deepening our understanding of the neuroscience underlying motor control, our study offers timely guidance for designing next-generation embodied intelligence systems. While our current framework focuses on online closed loop control, extending it to incorporate additional biological mechanisms such as memory consolidation and replay, as studied in lifelong reinforcement learning\cite{meng2025preserving}, offers a promising pathway toward more efficient, adaptive, and scalable learning in embodied systems. Such improvements in learning mechanisms are expected to influence the formation of control strategies and reshape the coordination between neural dynamics and body mechanics in embodied systems. We also observed that sensor and motor redundancy improves the robustness and generalizability of learning based control policies, and may help reshape current paradigms in robot design. In particular, as the development of humanoid robots and dexterous robotic hands increasingly relies on the coordination of large numbers of actuators, our study highlights how bio-inspired neural strategies, such as exploiting orthogonal latent dynamics and leveraging control redundancy, can help achieve adaptive, stable, and energy-efficient movements. These principles may also inform the development of responsive and naturalistic rehabilitation robots and prosthetic devices \cite{kilgard_closed-loop_2025, Nicolas2025Augmenting,Ophelia2021,Tony2023Understanding,Tony2024Understanding,Huang2025Grasping}. Ultimately, this research lays a foundation for translating principles of neural population dynamics into artificial control architectures, paving the way for more flexible and robust robotic systems.

\subsection*{Materials and Methods}

\subsubsection*{Disclosure of AI-Assisted technologies}
Reinforcement learning was used as a core method for training models in this study. This use falls within the scope of our specific research methodology. No large language models or other AI-assisted tools were used in the writing or presentation of this manuscript.

\subsubsection*{Fabrication of the physical model}

\textbf{Self-sensing artificial muscles.} The artificial muscles were fabricated by braiding LCE fibers coated with conductive fibers into a square knot structure \cite{Chen2025Scalable}. The fabrication process consists of three main steps. First, LCE fibers with a diameter of approximately 300~$\mu$m were printed via direct ink writing (DIW). The LCE ink was formulated by mixing 1,4-Bis-[4-(3-acryloyloxypropyloxy)benzoyloxy]-2-methylbenzene (RM257, Wilshire Technologies, 95\%), EDDET (Sigma-Aldrich, 95\%), dipropylamine (Sigma-Aldrich, 98\%), and (2-hydroxyethoxy)\\-2-methylpropiophenone (Irgacure 2959, Sigma-Aldrich, 98\%), using dichloromethane (Sinopharm, {CH2Cl2}) as the solvent. Specifically, RM257 (8.2404~g, 14~mmol) was dissolved in 50\,mL of dichloromethane. Subsequently, EDDET (2.1876~g, 12~mmol) and dipropylamine (0.100~g, 1~mmol) were added dropwise, and the mixture was stirred at room temperature overnight. A photoinitiator, Irgacure 651 (Medkoo, 0.0500~g, 0.2~mmol), was then incorporated. The resulting ink was placed in an oven at 100~\textdegree{}C for 24 hours to ensure complete solvent evaporation. The ink was then loaded into a syringe and mounted on a custom-built DIW 3D printer. Printing was carried out at a speed of 2~mm/s near room temperature (25~\textdegree C), with the flow rate controlled by maintaining an air pressure of 80~psi. After printing, the LCE fibers were carefully removed from the build plate and exposed to UV light (365~nm) for 1 hour to complete the crosslinking process. Second, we leveraged the centripetal force of a water vortex to coat 7~$\mu$m steel fibers onto the LCE fibers. The steel fibers form a furry network to make a conductive coating. Finally, we combined 2 conductive LCE fibers into a bundle and braided the bundle in a square knot structure with a knot distance of 6~mm to finalize the artificial muscle.

\textbf{IMU sensors.} The IMU sensors were designed with a master-slave architecture consisting of one master node and multiple slave nodes. The master node is an Arduino Nano 33 BLE board with a nRF52840 microcontroller. It is connected to all slave nodes via a software-based $I^2C$ bus and transmits the collected sensor data to a PC through the serial port for real-time processing and visualization. Each slave node, measuring 1.2~cm $\times$ 1.2~cm $\times$ 0.16~cm, is composed of a microcontroller unit (MCU, ATmega328P) and an IMU sensor (MPU6050), as shown in fig. S1. The MPU6050 sensor includes a 3-axis accelerometer and a 3-axis gyroscope, allowing the node to capture real-time orientation in quaternion format. 

\textbf{Rigid skeletal segment.}
The rigid skeletal segments adopt a triangular pyramid shape, which were fabricated using photopolymerization-based 3D printing (Formlabs Form 3, Formlabs Inc., USA) with a transparent resin. A latex membrane was attached to the bottom of each rigid skeletal segment to provide the necessary elastic restoring force for bending and contraction of the robot arm. Each rigid skeletal segment is connected to the central position of the latex membrane at the bottom of the preceding segment.

\subsubsection*{Assembly of the simulated model}
\textbf{Resistance-length relationship of the artificial muscle.} Based on the collected data on the resistance versus length under different electrical stimulation, we fit an empirical model relating resistance to length for the artificial muscles. Under different electrical excitations, the rate of change in both resistance and length followed the same exponential law:
\begin{equation}
\frac{\Delta L}{L_0} = l_1 e^{l_2\frac{\Delta R}{R_0}} + l_3,
\end{equation}
where $\Delta L$ denotes the length change of the artificial muscle, $L_0$ denotes the original length, $\Delta R$ denotes the resistance change of the artificial muscle, and $R_0$ denotes the original resistance. The constants $l_1$, $l_2$, and $l_3$ are fitted to be 0.0097, 5.1, and –0.0097, respectively (fig. S2B).

\textbf{Constitutive model of the artificial muscles.} To characterize the pure mechanical response of the artificial muscle, we adopted the Ogden hyperelastic model, as described below:

\begin{equation}
\sigma = \sum_{k = 1}^{3} \frac{2\mu_{k}}{\alpha_{k}} \left[ \lambda \alpha_{k} - \lambda^{\frac{\alpha_{k}}{2}} \right],
\end{equation}
where $\mu_{1}=0.5050$, $\alpha_{1}=7.5596$, $\mu_{2}=0.4360$, $\alpha_{2}=1.2870$, $\mu_{3}=-0.7117$, $\alpha_{3}=6.4458$. $\lambda$ is the ratio between the length of the artificial muscle and the original length of the artificial muscle (fig. S2E--F).

Since temperature is irrelevant to the force-strain profile, the thermomechanical model of the actuator extends the constitutive model (2) with an additional temperature-dependent term and a rate-dependent term \cite{Yip2017}:
\begin{equation}
\ F = a \sum_{k = 1}^{3} \frac{2\mu_{k}}{\alpha_{k}} \left[ \lambda \alpha_{k} - \lambda^{\frac{\alpha_{k}}{2}} \right] +b\dot{L}+c(T-T_0),
\end{equation}
where $a$ is the cross-sectional area of the artificial muscle, $b$ is the damping coefficient, and $\dot{L}$ is the rate of change in length. Damping parameter $b$ is found by dropping a 20g mass and measuring the decaying oscillation. The time constant of the decay is measured and is used to acquire a damping coefficient $b=1.043N\cdot s/m$ (fig. S2G). These parameters were obtained via least-squares fitting. Under different stimulation currents, temperature causes a linear shift in the force–strain profile (fig. S2H). The mean slope $c$ quantifying this effect is 0.0023N/\textdegree C.

Building upon our mechanical characterization, we developed a thermoelectric model to describe the energy conversion process during artificial muscle activation. When electric current charges the artificial muscle, we obtain:
\begin{equation}
\label{eq:thermoelectric-model}
C_{\mathrm{th}} \frac{dT(t)}{dt} = P(t) - q (T(t) - T_{\mathrm{amb}}),
\end{equation}
where $C_{\mathrm{th}}$ is the thermal mass of the artificial muscle (Ws/\textdegree C), $P(t)$ is the heat applied to the artificial muscle, q is the absolute thermal conductivity of the artificial muscle in the ambient environment (W/ \textdegree C), $T(t)$ is the temperature of the artificial muscle, and $T_{\mathrm{amb}}$ is the ambient temperature of the environment. $P(t)$ is controlled using current $I=\sqrt{P/R}$, and $R$ is the mean resistance of the artificial muscle. The $C_{\mathrm{th}}$ is calculated as 0.012175.

\textbf{Mechanical model of the latex membrane.} The mechanical model of the latex membrane is simplified to the combined effect of three linear elastic springs. Each spring is modeled as:
\begin{equation}
F=kx,
\end{equation}
where $F$ is the restoring force generated by the spring, $x$ is the extension from the equilibrium position, and $k$ is the spring constant that characterizes the membrane's stiffness, which is fitted to be 248mN/cm. The value of $k$ is determined by experimentally compressing two rigid skeletal segments together to stretch the latex membrane and obtained by minimizing the mean square error between the experimental data and the fitted spring model(fig. S1E).

\subsubsection*{Reinforcement learning model training}
\textbf{Control strategy.}
A reinforcement learning approach was adopted to learn control policies that achieve two objectives: (1) reaching arbitrary target positions, and (2) tracking continuous paths within the workspace. The control task was formalized as a Markov Decision Process (MDP), which models the sequential interaction between the agent (controller) and its environment (the robot arm). The MDP in our context comprised the following components:

\begin{itemize}
    \item \textbf{State space:}  The state vector $s_t$ captures the multi-modal sensory state of the robot arm at time step $t$, and is defined as:
    \begin{equation}
        \label{eq:state-space}
        s_t=\left[o_t^R, o_t^J, p_t, p_t^*\right],
    \end{equation}
    where $o_t^R$ denotes the normalized resistances of the artificial muscles, reflecting the internal actuation state; $o_t^J$ represents joint orientations encoded in quaternion format; $p_t$ is the current Cartesian position of the end effector; and $p_t^*$ is the desired target position.

    \item \textbf{Action space:} The action vector $a_t$ specifies the real-time current intensity applied to each of the 12 artificial muscles: 
    \begin{equation}
        a_t = \left[ I_t^j \mid j = 1, 2, \dots, 12 \right].
    \end{equation}
    Here, $I_t^j$ denotes the current applied to the $j$-th muscle. To prevent the actuator from being damaged by excessive current, $I_t^j$ was constrained to a maximum of 0.4A.

    \item \textbf{Reward function:} The reward function is designed to encourage minimizing the distance to the target position while penalizing excessive current usage. It consists of two components: a real-time feedback term $r_\text{acc}$, which penalizes tracking error and current usage, and a bonus term $r_\text{bonus}$, which provides a sparse reward upon successful task completion. The overall reward at time step $t$, denoted as $r_t$, is defined as follows:
    \begin{equation}
        r_t = r_\text{acc} + r_\text{bonus},
    \end{equation}
    \begin{equation}
        r_\text{acc} = -\left[(p_t-p^*_t)^T Q_e (p_t-p^*_t) + a_t^T R_a a_t\right].
    \end{equation}
    Here, $Q_e$ and $R_a$ are positive semi-definite weight matrices that penalize position error and current magnitude, respectively. An additional sparse reward $r_\text{bonus} = 100$ is granted whenever the Euclidean error $\|p_t - p^*_t\|_2$ falls below a precision threshold of 2 mm.
\end{itemize}

We employed the Soft Actor-Critic (SAC) algorithm~\cite{haarnoja2018soft}, a model-free reinforcement learning method known for sample efficiency and stability, to train the control policy entirely in simulation. Both the actor and critic networks shared the same architecture. It consisted of a linear input layer projecting to a 512-dimensional hidden representation, followed by an RNN layer based on 512-node LSTM units to capture temporal dependencies in sensorimotor feedback. A final linear output layer mapped the representation to a 12-dimensional action space corresponding to the artificial muscles. To accelerate the training process and improve data efficiency, we adopted a parallelized learning framework in which multiple simulation environments were executed concurrently for experience collection.

A unified training task was designed to encourage the policy to generalize across diverse initial states and target positions, promoting robustness and adaptability. Specifically, during each training episode, once the end effector successfully reached the current target, a new target position was randomly sampled within the workspace. The agent was incentivized to reach as many such targets as possible within the episode duration, thereby accumulating bonus rewards. This design ensured that the policy learned generalizable movement strategies rather than overfitting to specific trajectories or initial conditions.

To facilitate the transition of control policy from simulation to real-world implementation, we adopted the muscle dynamics randomization technique during training. This method randomizes key parameters of the artificial muscle during training to reflect real-world physical uncertainties and unmodeled dynamics. Specifically, a set of critical parameters governing the thermomechanical behavior of the artificial muscles was randomized across episodes. These parameters were sampled from predefined uniform distributions centered on experimentally measured values every training episode. In addition, Gaussian observation noise was added to exteroceptive feedback to imitate real-world sensor noise. Collectively, these domain randomization techniques improved the generalizability of the learned policy across variations in physical artificial muscles, sensor fidelity, and environmental conditions, ultimately enabling direct transfer from simulation to the physical environment without requiring fine-tuning or retraining.

\textbf{Different controller types.} To enable effective motion control of the robot arm under various sensory feedback, we trained neural network policies using various combinations of sensory inputs in simulation. The robot arm was equipped with multiple sensing modalities, including: (i) the Cartesian position of the end-effector, measured via the external motion capture system; (ii) the orientation of each limb segment, captured by the customized IMUs; and (iii) the real-time electrical resistance of each artificial muscle monitored through the control circuit, which actually reflects the relative length of the muscle. 

As described in Eq.~\eqref{eq:state-space}, we constructed the reinforcement learning state vector $s_t$ by combining different subsets of sensory feedback with the target position $p_t^*$. Specifically, we explored the following combinations of feedback modalities:
\begin{itemize}
    \item \textbf{Vision-only (Vis)}: consisting of the current end-effector position and the target position $\{p_t, p^*_t\}$.
    \item \textbf{Muscle length-only (Len)}: containing only the resistance-based feedback (muscle length) from artificial muscles $\{o^R_t, p^*_t\}$.
    \item \textbf{Muscle length plus joint angle (Len+Ang)}: consisting of resistance feedback (muscle lengths) and joint angles from IMUs $\{o^R_t, o^J_t, p^*_t\}$. 
    \item \textbf{Muscle length plus vision (Len+Vis)}: incorporating resistance feedback (muscle lengths) along with visual information $\{o^R_t, p_t, p^*_t\}$.
    \item \textbf{Joint angle plus vision (Ang+Vis)}: combining proprioceptive joint orientation $o^J_t$ with visual signals $\{o^J_t, p_t, p^*_t\}$.
    \item \textbf{Muscle length plus joint angle plus vision (Len+Ang+Vis)}: integrating all available modalities $\{o^R_t, o^J_t, p_t, p^*_t\}$.
\end{itemize} 

For controllers that rely exclusively on proprioceptive signals (e.g., resistance or joint angle), we introduced a predictor network to infer the missing visual information. This predicted visual feedback was then used as input to the corresponding controller policy. Two separate predictor models were developed for this purpose: one is trained using resistance feedback $o^R_t$ and the other is based on joint angle feedback $o^J_t$. These predictor-controller pipelines allowed us to evaluate the feasibility of vision-free control and assess the contribution of each sensing mode to the overall performance of the learned policy.

\textbf{Predictor network training.} To reconstruct the missing exteroceptive signal (i.e., end-effector position $p_t$) from proprioceptive signals ($o^R_t$, $o^J_t$), we trained data-driven predictor networks using supervised learning. These networks emulated the internal estimation mechanisms of biological systems, which relied heavily on proprioception in the absence of vision. Initially, predictors were trained in simulation using datasets collected from rollouts of a pre-trained control policy. This ensured that the training data reflected realistic trajectories and actuation patterns encountered during motor control. To enhance real-world applicability, the predictor was subsequently fine-tuned using a small set of real-world motion data collected from the physical robot arm. This adaptation phase accounted for sensor noise, actuation delays, and other imperfections inherent in the physical systems. 

The predictor network employed an architecture similar to the policy network, consisting of a linear input layer, an LSTM layer to capture temporal correlations in sequential sensor data, and a linear output layer. The network was trained to minimize the mean squared error (MSE) between predicted and ground-truth end-effector positions:
\begin{equation}
    \mathcal{L}_\text{obs} := \frac{1}{n} \sum_{p_t,\hat{p_t}\in\mathcal{D}} \left(p_t - \hat{p_t}\right)^2,
\end{equation}
where $\mathcal{D}$ denotes the training dataset, and $n$ is the number of training samples.

To enhance the robustness of the predictor under realistic control variations, we introduced stochasticity into the policy during data collection. Specifically, the controller’s action at each timestep was sampled from a Gaussian distribution:
\begin{equation}
	\tilde{a_t} \sim \mathcal{N} \left( a_t(s_t), \sigma^2(s_t) \right)
\end{equation}
where $a_t(s_t)$ and $\sigma(s_t)$ are the mean and standard deviation output by the SAC policy's actor network. This stochastic sampling better approximated the variability in real control executions and promoted predictor generalization.

\subsubsection*{Data acquisition}
\textbf{Primate data.} The neural activity and movement of monkeys were adopted from a publicly available primate dataset \cite{kalidindi_rotational_2021}. 

\textbf{Videography.} Six high-speed 26MP motion capture cameras (Mars26H; NOKOV Science \& Technology Co., Ltd., Beijing, China) were equipped with 8 mm lenses (Lens Basler 8 mm, C23-0824-5M) and placed surrounding the arena at regular intervals approximately 0.6m from the center. The cameras were stabilized using tripods mounted on a customized aluminum profile bracket, 
with dimensions of $60\,\mathrm{cm} \times 120\,\mathrm{cm} \times 60\,\mathrm{cm}$. All camera shutters were controlled synchronously by a 50 Hz Arduino hardware trigger by means of Phoenix Contact Sensor/Actuator cables. Images were transmitted via Basler USB 3.0 cables to an acquisition computer equipped with an AMD Ryzen 7 5800H CPU, an NVIDIA RTX 3050 GPU. We used the Campy camera acquisition software suite to encode videos from all cameras in real time.

\textbf{Calibration.} Cameras were calibrated using tools from NOKOV Science \& Technology Co., Ltd.. For intrinsic calibration, the software's calibration feature was utilized to independently estimate the parameters of each camera through the use of an L-shaped calibration rod. For extrinsic calibration, a T-shaped calibration wand was swept through the motion capture volume while all cameras simultaneously captured images. The software was then employed to detect the observation volume and calculate the rotation and translation vectors for each camera. Periodic calibration verification was conducted to ensure that the camera positions and orientations remained undisturbed during recording sessions. In practice, the system demonstrated high stability, with a single calibration remaining accurate for several months.

\textbf{Collecting sensory feedback.} Vision feedback was provided by a motion capture system continuously tracking the 3D position of an 8 mm reflective marker attached to the robot arm's end at 50 Hz, providing real-time visual input. Muscle length feedback was derived from resistance measurements using a resistance-length model. Resistance was measured by a custom control circuit with voltage feedback, sampled at 5 Hz, and calculated by dividing the circuit’s voltage signal by the controller’s output current command. Joint angle feedback came from an IMU sensor providing quaternion data at 50 Hz, which was then converted into angular measurements to obtain joint angles.

\subsubsection*{Data analysis}
All data analysis was performed in custom-designed procedures in Matlab (Mathworks, R2022b) or Python (https://www.python.org). 

\textbf{The jPCA Analysis.} We performed jPCA analysis on the neural network according to the methodology reported in \cite{churchland_neural_2012}, using the code available at https://churchland.zuckermaninstitute.c\\olumbia.edu/content/code. We constructed matrices $X$ that contained the activities of all neurons in the network for every time point and condition (i.e., load combinations or reach directions). These matrices had $N \times CT$ dimensions, where $N$ is the number of neurons in the network, $C$ is the number of conditions, and $T$ is the number of time points.

PCA was applied to $X$ to get the top 30 principal components, which are used to reduce $X$ to $X_{red}$. We numerically calculated the derivative of $X_{red}$ to yield $\dot{X}_{red}$, and fitted a linear dynamical model which found a relationship between $X_{red}$ and $\dot{X}_{red}$ as:
\begin{equation}
\dot{X}_{red}=M_{\mathrm{skew}}X_{red}
\end{equation}
where $M_{\mathrm{skew}}$ is $30\times30$ skew-symmetric transition matrix. 

\textbf{Quantifying rotational dynamics.} To quantitatively evaluate the presence of rotational neural dynamics in our sensorimotor control system, we computed the angle $q$ between the neural state vector $X$ and its temporal derivative $\dot{X}$, along with the goodness of fit ($R^2$) from fitting the population activity to a low-dimensional rotational dynamical system. $R^2$ was determined by assessing the jPCA model’s fit. Specifically, $R^2$ was computed by first projecting the high-dimensional neural data onto the top principal components 
and then fitting the resulting trajectories with a skew-symmetric matrix $M_{\mathrm{skew}}$ that defines the jPCA plane. 
The predicted neural state at time $t$ was obtained as $X_{\mathrm{pred}}(t) = M_{\mathrm{skew}} X(t)$, 
and $R^2$ was calculated as the proportion of variance in the observed temporal derivative $\dot{X}(t)$ 
explained by $X_{\mathrm{pred}}(t)$:
\begin{equation}
R^2 = 1 - \frac{\sum_t \| \dot{X}(t) - X_{\mathrm{pred}}(t) \|^2}{\sum_t \| \dot{X}(t) - \overline{\dot{X}} \|^2}.
\end{equation}
Here, $\overline{\dot{X}}$ denotes the average of $\dot{X}(t)$ over time.

\textbf{Tensor maximum entropy surrogate data analysis.} We tested our findings against the hypothesis that rotational neural dynamics are a byproduct of the tuning and smoothness properties of neurons. To do so, we employed TME to generate surrogate data sets \cite{Elsayed2017} using code available at https://github.com/gamaleldin/TME (Kalidindi, 2021; copy archived at swh:1:rev:ad1adf835e72db\\ba012406b5a3af30701adc8993). This method generates surrogate data sets that preserve the covariances across neurons, conditions, and time but not their interactions required for rotational dynamics. We then sampled surrogate data sets from this distribution and applied jPCA analysis to each data set (100 iterations).

\textbf{Reconstruction of rotational dynamics dominated movement.}
To extract rotational dynamics in neural activity, we applied jPCA, which begins by reducing the original data dimensionality via PCA and then fits a skew-symmetric matrix $\mathbf{M}{\text{skew}}$ to capture the dominant rotational patterns. Decomposing $\mathbf{M}{\text{skew}}$ yields complex conjugate eigenvectors that define planes of rotation. These are transformed into real-valued projection vectors ${\mathbf{u}_{i,1}, \mathbf{u}_{i,2}}$, which span the jPCA planes containing the strongest rotational structure.

We focused on the first projection vector $\mathbf{u}_{i,1}$ (corresponding to the jPC1 signal), which we term the Rotation state. To reconstruct this extracted rotational dynamic back into synthetic neural activity in the original neural space, we projected $\mathbf{u}_{i,1}^\top$ through the PCA loading matrix $\mathbf{L} \in \mathbb{R}^{30 \times 512}$, yielding:
\begin{equation}
\mathbf{s}_{\text{rot}} = \mathbf{u}_{i,1}^{\top} \mathbf{L}
\end{equation}

This vector captures the rotational mode across all 512 neurons in the actor network of RNN layer. We then replicated $\mathbf{s}_{\text{rot}}$ over 60 time steps to produce a time-invariant neural signal:
\begin{equation}
\mathbf{S}_{\text{rot}} = \mathrm{rep}(\mathbf{s}_{\text{rot}}, 60)
\end{equation}

This synthetic signal, termed the ``neural signal'', served as the current state of the RNN layer in the actor network. The actor network output a control command, denoted ``Action'', which was applied to the robot arm. The resulting movement, driven by the rotational dynamics, was referred to as ``jPC1-reconstructed'', representing a behavioral instantiation of the ``Rotation state''.

\textbf{Velocity of robot arm.} We calculated the robot arm's velocity as the straight - line distance from the starting point to the target point divided by the time required to reach the target.

\textbf{Aligning rotational dynamics through CCA.} To address the hypothesis that the rotational neural dynamics generating orthogonal adjustment to the reaching direction, we aligned the latent dynamics estimated from neural population recording to behavior output using canonical correlation analysis (CCA), which maximizes the correlations between neural and behavioral signals via linear transformations. The linear combinations of biological or artificial neural activity can be projected onto a two-dimensional subspace, in which neural trajectories are well-aligned with corresponding two-dimensional movement trajectories in physical space. We used the same linear transformations to project the synthetic neural signal derived from rotational dynamics into the behavioral 2D space (jPC1 reconstructed-CCA), to estimate the effect of rotational component. 

\textbf{Quantification of unit contributions to rotational dynamics.} 
To characterize the contribution of individual recurrent units to rotational neural dynamics, we performed a multi-step analysis combining dimensionality reduction, phase-coupling quantification, and structural perturbation.

Recurrent neural activity was first concatenated across trials and time and projected onto a low-dimensional subspace using PCA. jPCA was then applied to the leading principal components to identify dominant rotational planes. The first pair of jPC axes (jPC1 and jPC2) was used to define the primary rotational structure.

To quantify the contribution of individual recurrent units, we defined a score that combines spatial loading onto the rotational plane and temporal coupling to the rotational phase. For each unit \( u \), its projection onto the rotational plane was defined as:
\begin{equation}
L_u = \sqrt{v_1(u)^2 + v_2(u)^2}
\end{equation}
where \( v_1(u) \) and \( v_2(u) \) denote the loadings of unit \( u \) on jPC1 and jPC2, respectively.

We then quantified the coupling between each unit’s activity and the population-level rotational phase. The instantaneous phase of the population activity was defined as:
\begin{equation}
\theta(t) = \mathrm{atan2}(\mathrm{jPC2}(t), \mathrm{jPC1}(t))
\end{equation}
Each unit’s activity was regressed onto \( \cos\theta(t) \) and \( \sin\theta(t) \), and the magnitude of the resulting regression coefficients was used to define a phase-coupling measure \( C_u \).

The overall contribution score for each unit was then defined as:
\begin{equation}
S_u = L_u \cdot C_u
\end{equation}
Units were ranked according to \( S_u \), and those with the highest scores were identified as dominant contributors to rotational dynamics.

To assess the structural role of individual units, we quantified their connectivity to input and output layers. The input connectivity strength for unit \( u \) was defined as the L2 norm of its input weights:
\begin{equation}
I_u = \| W_{\mathrm{in}}(:, u) \|_2
\end{equation}
Similarly, the output connectivity strength was defined as the L2 norm of its output weights:
\begin{equation}
O_u = \| W_{\mathrm{out}}(u, :) \|_2
\end{equation}

\textbf{Pruning analysis.} 
To test the functional importance of units contributing strongly to rotational dynamics, we performed a targeted pruning analysis. Units with the highest contribution scores were selected, and their direct input and output connections were removed by setting the corresponding weights in \( W_{\mathrm{in}} \) and \( W_{\mathrm{out}} \) to zero. The modified network was then evaluated on the same control task without retraining. Control performance was quantified using task-specific metrics (e.g., tracking error or reward). The effect of pruning on rotational structure was assessed by reapplying the jPCA analysis to the recurrent activity of the modified network.

\textbf{Subspace construction and null-space variance analysis.} 
To quantify the relationship between preparatory and execution neural activity, we performed a subspace-based analysis using PCA, followed by explicit definition of null spaces and variance projection.

The network hidden activity was divided into preparatory (\( H_{\mathrm{prep}} \in \mathbb{R}^{T_p \times N} \)) and execution (\( H_{\mathrm{exec}} \in \mathbb{R}^{T_e \times N} \)) epochs based on task timing, where \( T_p \) and \( T_e \) denote the number of time points in each epoch, and \( N \) is the dimensionality of the neural (hidden) state. All analyses were performed on condition-averaged activity to improve signal-to-noise ratio and capture population-level structure.

We performed PCA separately on preparatory and execution activity and retained the top \( k_p \) and \( k_e \) principal components that explained 90\% of the variance, forming orthonormal bases. The preparatory subspace basis is denoted as \( B_{\mathrm{prep}} \in \mathbb{R}^{N \times k_p} \), and the execution subspace basis as \( B_{\mathrm{exec}} \in \mathbb{R}^{N \times k_e} \).

For a given subspace basis \( B \), the projection onto the subspace (potent space) is defined as:
\begin{equation}
P_{\mathrm{potent}} = B B^\top
\end{equation}
The corresponding null space (orthogonal complement) is defined as:
\begin{equation}
P_{\mathrm{null}} = I - B B^\top
\end{equation}
where \( I \) is the identity matrix.

For any activity matrix \( H \), the fraction of variance lying in the null space is computed as:
\begin{equation}
\mathrm{Var}_{\mathrm{null}}(H) = \frac{\| H P_{\mathrm{null}} \|_F^2}{\| H \|_F^2}
\end{equation}
where \( \| \cdot \|_F \) denotes the Frobenius norm.

Using this definition, we computed the fraction of execution variance lying in the preparatory null space:
\begin{equation}
\mathrm{Var}_{\mathrm{exec \ in \ prep\mbox{-}null}} = \frac{\| H_{\mathrm{exec}} (I - B_{\mathrm{prep}} B_{\mathrm{prep}}^\top) \|_F^2}{\| H_{\mathrm{exec}} \|_F^2}
\end{equation}
and the fraction of preparatory variance lying in the execution null space:
\begin{equation}
\mathrm{Var}_{\mathrm{prep \ in \ exec\mbox{-}null}} = \frac{\| H_{\mathrm{prep}} (I - B_{\mathrm{exec}} B_{\mathrm{exec}}^\top) \|_F^2}{\| H_{\mathrm{prep}} \|_F^2}
\end{equation}

\textbf{Subspace overlap.} 
To quantify similarity between preparatory and execution subspaces, we computed the subspace overlap as:
\begin{equation}
\mathrm{Overlap} = \frac{1}{k} \| B_{\mathrm{prep}}^\top B_{\mathrm{exec}} \|_F^2    
\end{equation}
where \( k = \min(k_p, k_e) \). This metric ranges from 0 (orthogonal subspaces) to 1 (identical subspaces).

\textbf{Local linearization of recurrent dynamics.} 
To characterize the dynamical properties of recurrent activity, we performed a local linearization analysis of the network dynamics.

Let \( h_t \in \mathbb{R}^N \) denote the network hidden state at time \( t \). Neural activity was divided into preparatory and execution epochs based on task timing. Unless otherwise specified, all analyses were performed on condition-averaged activity.

Because the analytical form of the recurrent dynamics was not directly accessible, we estimated an empirical Jacobian by fitting a first-order linear dynamical system to the observed hidden state transitions within each epoch. After dimensionality reduction using PCA, the dynamics were approximated as:
\begin{equation}
\tilde{h}_{t+1} \approx A \tilde{h}_t
\end{equation}
where \( \tilde{h}_t \) denotes the low-dimensional projection of the hidden state, and \( A \) is the empirical Jacobian estimated via least-squares regression. Separate Jacobians were computed for the preparatory and execution epochs.

The eigenspectra of the estimated Jacobians were computed. Eigenvalues near the real axis and close to unity correspond to slow or weakly decaying modes, while complex eigenvalues correspond to oscillatory modes.

To quantify the relationship between dynamical modes and output-relevant structure, the leading eigenmodes of the Jacobian were projected onto the output-potent and output-null subspaces defined from the execution-epoch readout. The relative energy of each mode within these subspaces was computed.

The local output sensitivity was quantified using the Jacobian of the hidden-to-output mapping. Let \( m_t = g(h_t) \) denote the network output. Around a representative state, the output Jacobian is defined as:
\begin{equation}
J_m = \frac{\partial g}{\partial h}
\end{equation}
When the output layer is linear, \( J_m \) reduces to the output weight matrix. For preparatory and execution subspaces, the output gain was computed as the squared Frobenius norm of the projected Jacobian.

\textbf{Neural manifold–kinematics alignment analysis.} 
To quantify the relationship between neural population activity and behavioral output, we analyzed the alignment between neural and kinematic manifolds.

Neural activity was projected onto a low-dimensional manifold using PCA. A common neural manifold was constructed by applying PCA to the full dataset and retaining the top \( d_n \) components (fixed to 30 dimensions across all conditions), yielding latent neural trajectories \( Z(t) \in \mathbb{R}^{d_n} \).

Behavioral output (muscle activity or kinematics) was similarly projected onto a low-dimensional manifold using PCA. The top \( d_k \) components explaining 90\% of the variance were retained, resulting in kinematic trajectories \( K(t) \in \mathbb{R}^{d_k} \).

To identify behaviorally relevant neural subspaces, CCA was applied between neural latent activity \( Z(t) \) and kinematic activity \( K(t) \) within each task epoch (preparatory and execution). CCA computes pairs of linear projections that maximize the correlation between the two datasets, yielding kinematics-aligned neural subspaces.

The strength of neural–kinematic alignment was quantified using cross-validated canonical correlations.

To assess the specificity of alignment, cross-epoch analyses were performed by applying the CCA mappings derived from one epoch to neural activity from the other epoch. The resulting canonical correlations were compared with those obtained under matched conditions.

\textbf{Mean shift analysis between preparatory and execution states.} 
To quantify the displacement of neural population activity between preparatory and execution epochs, we computed centroid distances at both the trial-wise and population levels.

For each trial, neural activity was divided into preparatory and execution epochs. The centroid of each epoch was defined as the mean neural state across time. The Euclidean distance between the preparatory and execution centroids was computed for each trial. The mean and standard deviation of these distances across trials were then calculated.

Neural activity from preparatory and execution epochs was aggregated across all trials. Global centroids were computed for each epoch as the mean neural state over all samples. The Euclidean distance between these global centroids was used to quantify the overall mean shift.

Statistical significance was assessed using permutation tests. Within each trial, epoch labels were randomly shuffled while preserving the number of samples in each epoch. Centroid distances were recomputed for each permutation to generate a null distribution. The observed distances were compared with this null distribution to compute permutation-based \( p \)-values and \( z \)-scores.

\textbf{Functional decomposition of neural population activity.} 
To characterize distinct components of neural population activity, we decomposed neural activity into rotational, output-potent, and condition-invariant components within a common low-dimensional space.

The output-potent subspace was defined by identifying neural dimensions that best predict behavioral variables. A linear regression model was fit from neural activity to the target error vector using training data. Singular value decomposition (SVD) of the regression weight matrix was performed, and the leading singular vectors were retained to define the output-potent subspace.

The condition-invariant (CI) component was defined as the leading principal component of the across-condition mean neural activity over time.

To quantify the relationship between neural components and behavioral variables, \( K \)-fold cross-validation was performed across trials. Within each fold, regression models were trained on the training set and evaluated on held-out data to predict the target error vector and target-directed velocity. Predictive performance was quantified using the coefficient of determination \( R^2 \).

To quantify geometric relationships between subspaces, principal angles between the output-potent and jPC subspaces were computed using singular value decomposition of their basis matrices.

\textbf{Energy consumption.} To compute the neural energy for a single task, we extracted the latent activities of the 512 neurons from the RNN during the robotic analog’s reaching movement. Let \( a_i(t) \) denote the latent activity of neuron \( i \) at time point \( t \), and let \( T \) be the total number of time points in the task. The neural energy \( E \) is then defined as:
\begin{equation}
E = \frac{1}{T} \sum_{t=1}^{T} \sum_{i=1}^{512} a_i(t)^2
\end{equation}

This metric provides a quantitative measure of the overall neural activation level associated with a single task.

Additionally, we calculated the average energy consumption of each muscle during the reach task using the formula \( W = I^2 R t \), where \( I \) is the current, \( R \) the resistance, and \( t \) the time duration. This provided a measure of muscle energy expenditure for each task, allowing us to assess the energy efficiency of different movement strategies. Examining the relationship between neural activity and muscle energy enables us to investigate how various sensory feedback configurations influence the energy demands of both the neural controller and the muscles during movement.

\textbf{The normalized radius of neural rotation.} We quantified rotational neural dynamics by fitting neural trajectories with smooth parametric curves in the neural state space and computing their radius of curvature. These values were then normalized to obtain a standardized radius of neural rotation, providing a consistent measure of rotational strength across different conditions.

\textbf{PCA variance explained.} To assess how neural population dynamics evolve during training, we computed the variance explained by principal components (PCA variance explained) at different training stages. Specifically, at selected training episodes, we concatenated neural activity data across 20 trials and all time points during the reaching tasks, organizing the data into a single matrix where rows correspond to time points across all trials and columns correspond to neural units. PCA was then applied to this combined data matrix to extract principal components and calculate the variance explained by each component. By tracking the variance explained by the top two principal components over the course of training, we evaluated how the dimensional structure of neural population dynamics changed as learning progressed.

\textbf{The Angular separation.} To quantify the difference between preparatory and execution neural states, we computed the angle $\xi$ between the first principal components of the PCA subspaces corresponding to these two periods.

Specifically, for each trial, we first performed PCA to reduce the 512-dimensional neural activity data (size $60 \times 512$) to a 30-dimensional space. The first 5 time points (covering preparatory state) and the remaining 55 time points (covering execution state) were then treated as separate datasets in this reduced space.

We applied SVD to each dataset to obtain orthonormal bases, denoted as $\mathbf{V}_{\mathrm{prep}}$ and $\mathbf{V}_{\mathrm{exec}}$. From each, we extracted the first right singular vector (i.e., the first principal component) $\mathbf{v}_{\mathrm{prep}}$ and $\mathbf{v}_{\mathrm{exec}}$. The angle $\xi$ between these vectors was calculated as:
\begin{equation}
\xi = \arccos \left( \frac{\left| \mathbf{v}_{\mathrm{prep}} \cdot \mathbf{v}_{\mathrm{exec}} \right|}{\|\mathbf{v}_{\mathrm{prep}}\| \, \|\mathbf{v}_{\mathrm{exec}}\|} \right)
\end{equation}
where the dot product was clipped to $[-1, 1]$ for numerical stability, and the resulting angle was converted from radians to degrees.

This angle $\xi$ quantifies how distinct the preparatory and execution PCA subspaces are: larger angles indicate more orthogonal (i.e., more distinct) neural representations between preparation and movement execution states.

\subsection*{Supplementary materials}
Materials and Methods\\
Supplementary Text\\
Fig. S1 to S19\\
References \textit{(1-7)}\\ % automatically fills out the last reference number
% (filling out the other numbers automatically is possible but fiddly and liable to break)
Movie S1 to S4\\

%%%%%%%%%%%%%%%% REFERENCES %%%%%%%%%%%%%%%

\clearpage % Clear all remaining figures and tables then start a new page

% The list of references goes after the main text and before the acknowledgements
% When preparing an initial submission, we recommend you use BibTeX, like this:
%

\bibliography{science_template} % for a file named science_template.bib
\bibliographystyle{sciencemag}

% After the paper has completed peer review and been revised ready for acceptance,
% you should comment out the lines above and copy-paste the contents of your .bbl
% file here instead. This will help ensure that our conversion software works correctly.
% Remember to re-run BibTeX first - check the timestamp!
%
% Example of the first three entries copy-pasted from science_template.bbl:
%
%\begin{thebibliography}{1}
%
%\bibitem{example}
%A.~N. {Author}, An example reference. \emph{Journal of Improbable Research}
%  \textbf{1}, 67 (2020).
%
%\bibitem{example2}
%F.~M. {Surname}, S.~{Author}, A second example. \emph{Interesting Research
%  Letters} \textbf{32}, 897 (2019).
%
%\bibitem{example_preprint}
%P.~{One}, P.~{Two}, P.~{Three}, {An unpublished preprint}. \emph{preprint}
%  (2021), arXiv:2101.12345.
%
%\end{thebibliography}

%%%%%%%%%%%%%%%% ACKNOWLEDGEMENTS %%%%%%%%%%%%%%%

\section*{Acknowledgments}
We thank Ms. Yiwei Hu for suggestions on the visualization.
\paragraph*{Funding:}
This research is supported by the National Key Research and Development Program of China through grant 2022YFB4701900, the National Natural Science Foundation of China through grant 12372159, the Fundamental Research Funds for the Central Universities, Peking University, the State Key Laboratory of Membrane Biology, Peking University, and the Peking-Tsinghua Center for Life Sciences.
\paragraph*{Author contributions:}
Conceptualization, K.L., X.L., F.Z., W.C.; data curation, W.C. and J.T.;  formal analysis, W.C. and J.T.; investigation, W.C., J.T., Y.C. and Y.S.; methodology, W.C. and J.T.; project administration, K.L.; resources, K.L.; software, W.C. and J.T.; supervision, K.L., X.L., F.Z.; validation, W.C. and J.T.;  visualization, W.C., K.L. and X.L.; writing—original draft, W.C., J.T., K.L., X.L.; and writing—review and editing, all authors.
\paragraph*{Competing interests:}
There are no competing interests to declare.
\paragraph*{Data and materials availability:}
All data is included in the main text, supplementary materials, and the Github repository for code deposition (https://github.com/chriskkkkl/Neural-dynamics-of-the-robotic-twin).

%%%%%%%%%%%%%%%% SUPPLEMENT LIST %%%%%%%%%%%%%%%

% List the contents of your Supplementary Materials, including the numbers of any
% supplementary figures, tables, external data files etc. and any references that are
% cited only in the supplement. In this example, refs. 7-8 are cited only in the supplement.
% Fill out your numbers accordingly and delete any lines that aren't applicable.

%%%%%%%%%%%%%%%% END OF MAIN TEXT %%%%%%%%%%%%%%%

\newpage

%%%%%%%%%%%%%%%% START OF SUPPLEMENT %%%%%%%%%%%%%%%

% Figures, tables, equations and pages in the supplement are numbered S1, S2 etc.
\renewcommand{\thefigure}{S\arabic{figure}}
\renewcommand{\thetable}{S\arabic{table}}
\renewcommand{\theequation}{S\arabic{equation}}
\renewcommand{\thepage}{S\arabic{page}}
\setcounter{figure}{0}
\setcounter{table}{0}
\setcounter{equation}{0}
\setcounter{page}{1} % not 0 as \newpage already started a supplementary page

\end{document}